\pdfoutput=1

\documentclass[11pt]{article}

\usepackage[preprint]{acl}

\usepackage{times}
\usepackage{latexsym}

\usepackage[T1]{fontenc}

\usepackage[utf8]{inputenc}

\usepackage{microtype}

\usepackage{inconsolata}

\usepackage{graphicx}
\usepackage{rotating}
\usepackage{adjustbox}
\usepackage{multirow}
\usepackage{arydshln}
\usepackage{amsthm}
\theoremstyle{plain}

\usepackage{amsmath,amssymb}
\usepackage{booktabs}
\usepackage[table]{xcolor}
\usepackage{multirow}
\usepackage[most]{tcolorbox}
\usepackage{xcolor}
\usepackage{listings}
\usepackage{subcaption}
\usepackage{makecell}
\usepackage{enumitem}

\newtcblisting{graycodebox}{
  listing only,
  colback=gray!6,
  colframe=gray!40,
  boxrule=0.6pt,
  arc=2pt,
  left=6pt,right=6pt,top=6pt,bottom=6pt,
  listing options={
    language=Python,
    basicstyle=\ttfamily\footnotesize,
    columns=fullflexible,
    breaklines=true,
    breakatwhitespace=true,
    showstringspaces=false,
    tabsize=2,
  }
}

\usepackage{pifont}
\newcommand{\cmark}{\ding{51}}
\newcommand{\xmark}{\ding{55}}

\title{StitchRAG: Constructing Compact and Evidence-Preserving Contexts for RAG with Diffusion-Based Stitching}

\author{
 \textbf{Minsang Kim\textsuperscript{1,2}},
 \textbf{Junpyo Lim\textsuperscript{1}},
 \textbf{Heeje Cho\textsuperscript{1}},
 \textbf{Seungjun Baek\textsuperscript{1}}\thanks{Corresonding Author}
\\
 \textsuperscript{1}Korea University,
 \textsuperscript{2}SK Telecom,
\\
 \texttt{kmswin1@korea.ac.kr}
}

\begin{document}
\maketitle

\begin{abstract}

Retrieval-augmented generation (RAG) improves factuality by grounding large language
models (LLMs) in external evidence. However, long retrieved contexts incur computational overhead and can even hurt performance
due to noisy or irrelevant content. We propose \textsc{StitchRAG}, a framework for constructing compact context in RAG. \textsc{StitchRAG}
builds a faithful, information-dense context while dynamically compressing the retrieved passages. We adapt the size reduction ratio to each passage by formulating a utility maximization trading-off the retained information and computation costs. 
We then extract answer candidates and query-related keywords from passages as anchors and collect passage sentences containing anchors under the derived size constraint. Finally, we connect or stitch the collected sentences together using text diffusion to form a coherent and natural context. Experiments on five open-domain QA benchmarks show that
\textsc{StitchRAG} consistently outperforms diverse
multi-step and compression baselines. Extensive ablations show the effectiveness of dynamic size allocation, anchor preservation, and diffusion-based stitching.
\end{abstract}

\section{Introduction}\label{sec:intro}

Large language models (LLMs)~\cite{achiam2023gpt,huang2024llama} have rapidly advanced and enabled a wide range of real-world applications. However, they face fundamental limitations: updating their knowledge is costly, and LLMs often hallucinate responses that are plausible but incorrect~\cite{welleck2019neural, kalai2025language}. Retrieval-augmented generation (RAG) addresses these issues by grounding generation on external evidence retrieved at inference time~\cite{chen2017reading,lewis2020retrieval}. By conditioning on retrieved passages, RAG improves factuality and enables easy updates with up-to-date information.

\begin{figure}[t]
    \centering   \includegraphics[width=0.5\textwidth]{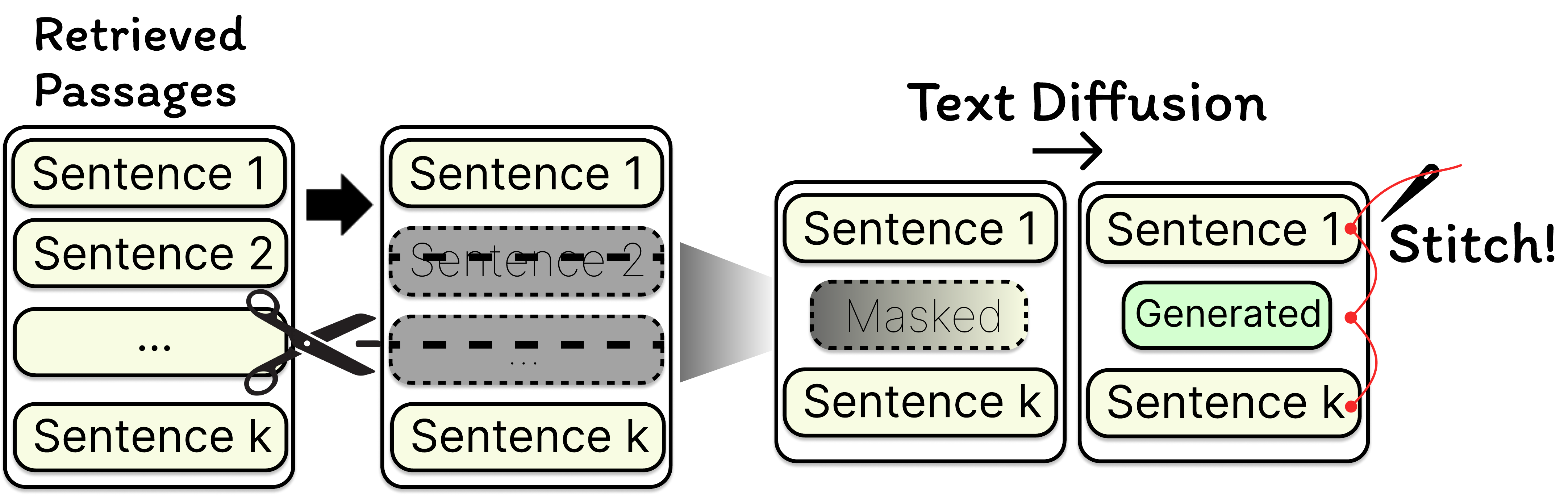}
    \caption{\textbf{Constructing Compact Context.} From retrieved passages, StitchRAG extracts and preserves sentences containing key terms. Then it connects them via diffusion-based stitching to construct a compact context.}
    \label{fig:example}
\end{figure}

A growing body of work improves RAG via {advanced retrieval and generation loops}, including multi-step reasoning~\cite{trivedi2023interleaving, kim2024qpaug}, self-reflection~\cite{jiang2023active, asai2023self}, and multi-agent debate or verification~\cite{du2024improving, hu2025removal}, in order to reduce hallucinations and strengthen reasoning. However, these methods primarily focus on iterative retrieval or post hoc verification, while largely leaving the context constructed from retrieved passages unconstrained. As a result, the context often accumulates redundant or conflicting evidence.
Such longer or repeatedly processed contexts also increase inference latency and computational cost. More importantly, even with strong retrievers, simply providing more evidence does not reliably improve answers: LLMs can be confused by noisy passages and long inputs~\cite{liu2024lost,hong2025context,du2025context}.

This has motivated substantial interest in \emph{context engineering}~\cite{mei2025survey, hua2025context} for RAG, which helps models better interpret the retrieved contexts. Context compression methods, such as selection~\cite{li2023compressing} and compression prior to generation~\cite{pan2024llmlingua, xu2023recomp}, reduce the inference cost and latency by shortening the input length. However, most existing approaches rely on fixed or heuristic compression ratios, largely independent of the query, passage quality, or reasoning complexity. In practice, a good compression should be context-dependent: some questions require preserving long, cross-passage evidence chains, whereas others benefit from aggressively filtering irrelevant content. Consequently, applying a single global compression ratio to all passages can create a mismatch between the context reduction and the information needs of each problem. 
Moreover, because prior compression methods typically operate at the token level rather than preserving original sentences, they often degrade the readability of contexts and thus the overall performance of RAG.

\noindent\textbf{Contribution.} We propose \textsc{StitchRAG}, a framework for constructing compact contexts for RAG. \textsc{StitchRAG} dynamically retains evidence in retrieved passages for size reduction and reconstructs a coherent context from preserved evidence. We formulate evidence retention as a constrained optimization problem: we maximize relevance utility combined with a computation-aware penalty to derive a closed-form retention ratio for each retrieved passage. This enables a dynamic adjustment of the retention ratio to each passage, improving the efficiency and relevance of the constructed context.
To build a compact yet readable context, we introduce hybrid anchor selection and diffusion-based stitching. Anchors are defined as key terms for the query, and sentences containing anchors are preserved in the context as fixed evidence. Next, we perform \emph{context stitching}, which connects fragmented sentences using text diffusion in order to restore coherence of the generated context. The proposed refinement enables faithful context reconstruction while reducing factual drift and hallucination. In our experiments, \textsc{StitchRAG} consistently improves the RAG performance over baseline methods. Extensive ablations confirm the importance of dynamic context retention, anchor preservation, and diffusion-based stitching.

\section{Related Work}\label{sec:related}

\noindent\textbf{Advanced RAG and Multi-step Retrieval/Generation.}
Recent work has explored \textit{multi-step} and \textit{adaptive} strategies that couple retrieval with intermediate reasoning steps. IRCoT~\cite{trivedi2023interleaving} interleaves reasoning traces with retrieval, enabling sequential evidence acquisition for multi-hop questions. FLARE~\cite{jiang2023active} triggers retrieval based on token-level uncertainty during generation, while Self-RAG~\cite{asai2023self} incorporates self-reflection to assess retrieved evidence and adapt the generation accordingly. SuRe~\cite{kim2024sure} is a multi-stage framework that generates answer candidates and summarizes retrieved passages conditioned on each candidate, then ranks candidates by verifying the resulting summaries. More recently, DRAG~\cite{hu2025removal} introduces a debate-based verification framework to reduce hallucinations and improve answer reliability under retrieval constraints.
However, these methods primarily focus on when to retrieve and how to expand or revise evidence through multi-step cycles. In contrast, StitchRAG addresses how to compress retrieved context under a dynamic budget while preserving crucial information and connecting fragmented evidence. 

\noindent\textbf{Context Compression.} 
Prior work has studied context compression as a means of reducing input length before generation. Selective Context~\cite{li2023compressing} prunes tokens using information-theoretic criteria (e.g., self-information), aiming to remove redundant content. LLMLingua and LongLLMLingua~\cite{jiang2023llmlingua,jiang2024longllmlingua} propose coarse-to-fine compression using a small language model to budget and select query-relevant spans. 
RECOMP~\cite{xu2023recomp} trains compressive models to distill retrieved documents into compact representations or summaries. Recent work also studies attention-guided pruning for RAG contexts~\cite{fang2025attentionrag}, which leverages attention signals to identify and remove redundant evidence. Many prior methods frame context compression as a separate pre-processing step, relying on static heuristics or auxiliary models. In contrast, StitchRAG incorporates budget allocation and evidence retention in context construction for RAG.

\noindent\textbf{Diffusion Models for Text Generation.}
Diffusion models provide a non-autoregressive framework for text generation via iterative refinement. Early works such as Diffusion-LM~\cite{li2022diffusion} and DiffuSeq~\cite{gong2022diffuseq} adapted denoising processes to discrete sequence modeling. Recent scalable approaches~\cite{nie2025large,ye2025dream} operate on masked generative principles and achieve performance competitive with autoregressive LLMs at scale. Their iterative refinement supports global editing with bidirectional context, which is particularly suitable for controlled text modification. Building on these properties, StitchRAG adopts a retrieve-and-refine paradigm.

\begin{figure*}
    \centering
    \includegraphics[width=1.\linewidth]{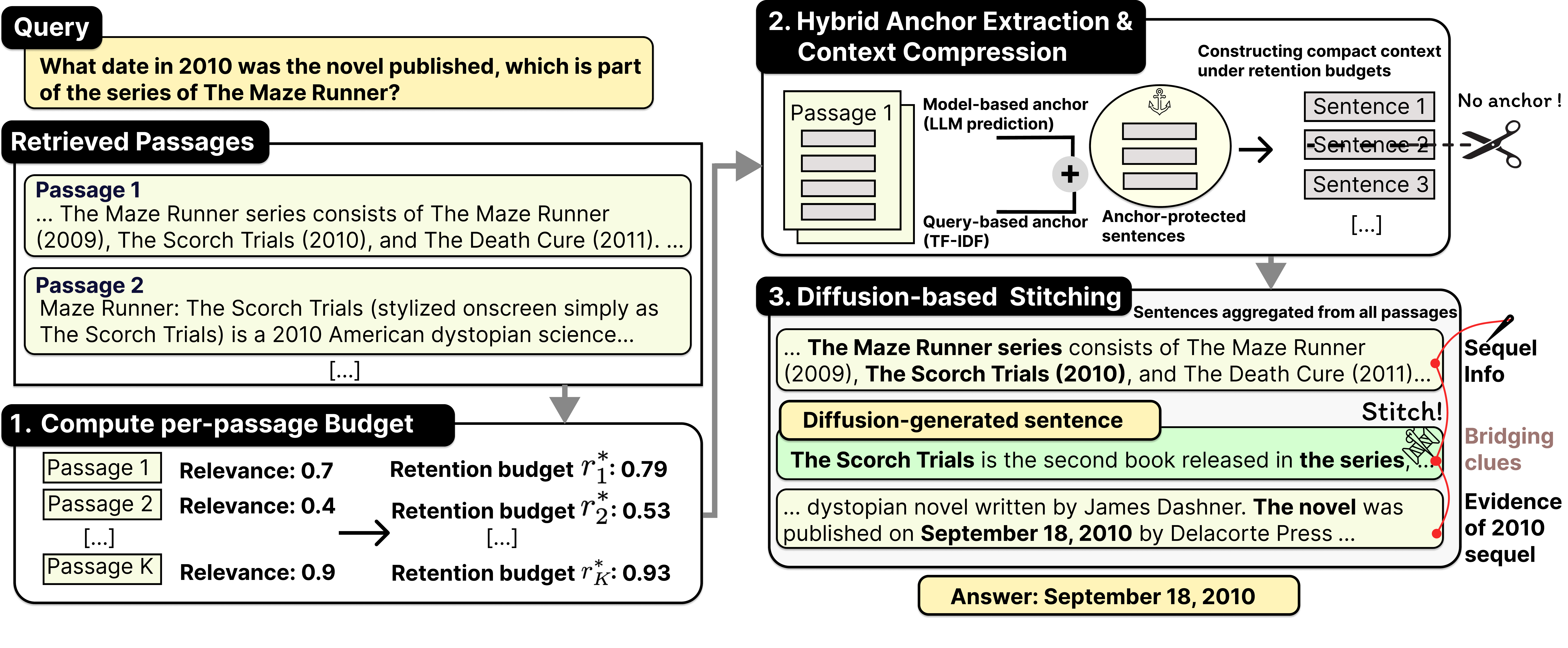}
    \caption{\textbf{Overview of StitchRAG.} Given a query and a set of retrieved passages, StitchRAG \textbf{1)} computes a per-passage dynamic retention ratio $r^*$, \textbf{2)} extracts passage sentences containing key terms, called \emph{anchors}, and aggregates them under the token budget constrained by $r^*$, and \textbf{3)} 
    generates bridging text between the aggregated sentences using a diffusion LLM. Finally, the constructed context is fed into the LLM to generate the answer.}
    \label{fig:method}
\end{figure*}

\section{Method}\label{sec:method}
StitchRAG is a framework to construct a compact context from retrieved passages for RAG. It first computes how much to reduce each passage in its size and determines its context budget; next, it retains only sentences containing key terms under the budget; finally, it bridges retained sentences via text diffusion. Fig.\ \ref{fig:method}. presents an overview.
\subsection{Optimal Context Retention via Utility Maximization}\label{sec:context_selection}

\noindent\textbf{Problem Formulation.}
We consider a standard RAG setting where a query $q$ retrieves $K$ passages $\mathcal{P}=\{x_1,\dots,x_K\}$. Consider compressing passage $x\in\mathcal{P}$ where the compression rate depends on $x$. For passage $x$, we define the \emph{retention ratio} $r(x,q)\in[0,1]$ (for simplicity, hereafter we drop the dependence of $r$ on $x,q$). The compressed passage has length $r\cdot |x|$ where $|x|$ denotes the length of passage $x$ in the number of tokens. For example, if $r=0.7$, the compressed passage has 70\% of the original length. Our goal is to find the optimal $r^*$ for given passage $x$. We formulate this problem as \emph{cost-aware utility maximization}. Specifically, we maximize the objective $\mathcal{F}(r)$ for $r\in[0,1]$:
\begin{align}
\mathcal{F}(r) = \mathcal{U}(r)-\mathcal{C}(r)    
\end{align}
where $\mathcal{U}$ denotes the achievable utility and $\mathcal{C}$ denotes the computation cost.


\noindent\textbf{Relevance utility.} The utility is defined to measure how the compressed passage is relevant to query $q$. However, in the first place, the original passage $x$ should be relevant to $q$ prior to compression. Suppose we obtain utility $u(r)$ with retention ratio $r$ if $x$ is relevant, and utility 0 if $x$ is irrelevant. Let random variable $Z\in\{0,1\}$ indicate relevance ($Z{=}1$ iff $x$ is relevant). The achievable utility is given by $Z\cdot u(r)$, and $\mathcal{U}(r)$ is defined as its expectation:
\begin{equation}
   \mathcal{U}(r) = \mathbb{E}[Z]\cdot u(r)= P(Z{=}1)\cdot u(r).
    \label{eq:relevance}
\end{equation}
We estimate $P(Z{=}1)$ by probing the LLM with a binary prompt that asks whether passage $x$ is useful to answer $q$ (see Appendix~\ref{appendix:prompts} for the template). We compute the log-probabilities of the LLM of the answers (``Yes'' vs.\ ``No''), denoted by $\mathcal{L}_{\text{Yes}}$ and $\mathcal{L}_{\text{No}}$, and convert them into a relevance score:
\begin{equation}    P(Z{=}1)\approx R(x,q):=\frac{\exp(\mathcal{L}_{\text{Yes}})}{\exp(\mathcal{L}_{\text{Yes}})+\exp(\mathcal{L}_{\text{No}})}.\nonumber
\end{equation}

Next, we define $u(r)$. Since $u(r)$ is obtained assuming $x$ is a relevant passage, $u(r)$ should be increasing in $r$; retaining more from $x$ is more informative to answer $q$. However, we assume a \emph{diminishing return}: new information does not increase proportionately with the context length. Thus, we adopt a \emph{logarithmic utility} in $r$:
\begin{align}
u(r) = \log(1+r)\label{eq:logutil}
\end{align}
 A diminishing return is consistent with Heaps' law~\cite{heaps1978information}, which states that the number of distinct words grows sub-linearly in the document length. Thus, redundancy grows faster than novel information as the context gets longer. Later, we give more justification for log-utility $u(r)$. In summary, $\mathcal{U}(r)=R(x,q)\log(1+r)$.

\noindent\textbf{Computation Cost.}
The latency of self-attention scales quadratically with the sequence length. Thus, the computation cost of retaining $r|x|$ tokens is proportional to $(r|x|)^2$. In RAG, documents are commonly segmented into passages of fixed word count~\cite{karpukhin2020dense} or fixed token length~\cite{bhat2025rethinking, zhong2025mix} for indexing. 
Thus, we treat $|x|$ as a constant and define the cost objective:
\begin{equation}
    \mathcal{C}(r)=\beta r^2,
    \label{eq:penalty}
\end{equation}
where the hyperparameter $\beta>0$ controls the latency penalty and absorbs the constant factor $|x|^2$.

\noindent\textbf{Deriving the optimal $r^*$.}
We maximize expected utility minus cost:
\begin{equation}
    \max_{r\in[0,1]} \ \mathcal{F}(r)=R(x,q)\cdot \log(1+r)-\beta r^2.\label{eq:opt}
\end{equation}
$\mathcal{F}(r)$ is a strictly concave function in $r$, and \eqref{eq:opt} admits a unique solution \cite{boyd2004convex} as follows:
\begin{equation}
r^*=\min\left\{\frac{1}{2}\left(\sqrt{1+\frac{2R(x,q)}{\beta}}-1\right),1\right\}.
    \label{eq:r_hat}
\end{equation}
We set hyperparameter $\beta = 0.25$. In this setting, one can show that $r^*=1$ if and only if $R(x,q)=1$, which is a reasonable choice. In \eqref{eq:r_hat}, $r^*$ increases with relevance confidence $R$, i.e., highly relevant passages should retain more context. 

\subsection{Hybrid Anchor Extraction}
\label{sec:anchor_extraction}
For passage $x$, our token budget is set to $r^*|x|$. We select and retain important \emph{sentences} from $x$ under this budget. Retaining sentences, instead of individual tokens, helps to maintain the coherence of context after compression. To that end, we select sentences which contain important spans, called \emph{anchors}. We first select the anchors by a hybrid strategy: a semantic-lexical selection that combines reasoning-centric generation with query-centric lexical matching.

\noindent\textbf{Reasoning-centric anchors (AR-based).} We generate multiple answer candidates using the LLM conditioned on $(q, x)$. The answer spans form the anchor set, $A^{\text{AR}}$, to capture implicit semantic dependencies beyond simple keyword matching. 

\noindent\textbf{Query-centric anchors (TF-IDF-based).}
In parallel, we extract the query-centric anchor set denoted by $A^{\text{TFIDF}}$ using a TF-IDF score to ensure the coverage of rare entities and lexical cues. The final anchor set is the union of the anchor sets:
\begin{equation}
    \mathcal{A}=A^{\text{AR}}\cup A^{\text{TFIDF}}.
\end{equation}
\subsection{Fact-First Sentence Selection}
\label{sec:sentence_selection}
Given the token budget $T_{\text{budget}} = r^* \cdot |x|$, we select sentences from $x$ that contain anchors. We initially mark all the sentences in $x$ as ``unselected''.
\noindent\textbf{1. Anchor protection.}
We identify sentences that contain anchors in $\mathcal{A}$ and mark them as ``selected''. Let $\mathcal{S}_{\mathcal{A}}$ denote this set of sentences, which forms the factual backbone of the compressed context.
\noindent\textbf{2. Budget-aware selection.}
If the total number of tokens in $\mathcal{S}_{\mathcal{A}}$ is less than $T_{\text{budget}}$,
we select some non-anchor sentences using a lightweight priority rule: (i) sentences containing query keywords, and (ii) earlier sentences in retrieved passages, which often summarize salient local context. These high-priority non-anchor sentences are marked as ``selected'' until the token budget is reached. Finally, all the sentences marked as ``unselected'' are removed from $x$ to obtain the compressed passage.

\noindent\textbf{Passage Compression and Log-utility.} We provide a heuristic argument relating log-utility $u(r)$ to our passage compression. Suppose we randomly choose a sentence from passage $x$. The probability of choosing an anchor sentence is \(p=\frac{|\mathcal{S}_{\mathcal{A}}|}{|\mathcal{S}|}\). If we take the logarithm of $1/p$, we have that
\begin{align}
\log \frac{1}{p} = \log \frac{|\mathcal{S}|}{|\mathcal{S}_{\mathcal{A}}|}\label{eq:idf}
\end{align}
\eqref{eq:idf} is analogous to the well-known \emph{Inverse Document Frequency (IDF)}; consider $\mathcal S$ as a collection of documents (treating sentences as documents), and $\mathcal{S}_{\mathcal{A}}$ as the document set containing a keyword, then \eqref{eq:idf} is the IDF. The IDF serves as a relevance metric for retrieval \cite{robertson2004understanding}. Suppose $x$ is compressed by our sentence selection method with retention ratio $r$. The compressed passage retains $\mathcal{S}_{\mathcal{A}}$, but its size is reduced to approximately $r|\mathcal{S}|$. Thus, \eqref{eq:idf} for the compressed passage is
\begin{align}
\log \frac{r|\mathcal{S}|}{|\mathcal{S}_{\mathcal{A}}|} = \log r + \log \frac{|\mathcal{S}|}{|\mathcal{S}_{\mathcal{A}}|}\label{eq:idf-comp}
\end{align}
\eqref{eq:idf-comp} shows the logarithmic dependence of relevance metric on $r$ (the second term is constant and does not affect optimization \eqref{eq:opt}), where we used $u(r)=\log(1+r)$ for numerical stability. Overall, $u(r)$ can be approximately seen as a relevance metric associating the compressed passage with anchor terms.

\subsection{Diffusion-based Context Stitching}
\label{sec:stitching} All the retrieved passages are compressed and then concatenated to construct the final context denoted by $\mathcal{X}$. While $\mathcal{X}$ preserves crucial factual information,  concatenated sentences may not form a coherent context as a whole, because (i) some sentences are removed within the passage during compression; (ii) possibly heterogeneous passages are directly concatenated. We propose \emph{context-stitching} based on text diffusion. The goal of stitching is to simply connect the sentences to maintain their natural semantic flows using a small amount of tokens. The stitching consists of two steps:

\begin{enumerate}[topsep=2pt, itemsep=2pt, parsep=0pt, partopsep=0pt]
    \item Dynamically allocate \emph{stitching tokens}, initialized as masked tokens, and insert them between adjacent sentences in $\mathcal{X}$.
    \item Fill these stitching tokens in parallel via a diffusion-based denoising process.
\end{enumerate}

\noindent\textbf{Allocation based on Semantic overlap.}
We allocate $\kappa_{i,j}$ stitching tokens between sentences $i$ and $j$ based on their semantic overlap. Let $\mathcal{J}_{i,j}$ denote the Jaccard similarity of sentences $i$ and $j$, then 
\begin{equation}    \kappa_{i,j}=\left\lfloor\bar\kappa\cdot \bigl(1-\mathcal{J}_{i,j}\bigr)\right\rfloor\label{eq:sim}
\end{equation}
where $\bar\kappa$ is the maximum budget for each stitching. $\bar\kappa$ may differ depending on the types of $i,j$: if $i$ and $j$ belong to different passages, i.e., they are at passage boundaries and are likely to have a large semantic distance, $\bar\kappa$ is set relatively large. In contrast, if $i$ and $j$ belong to the same passage, $\bar\kappa$ is set relatively small; see Sec.\ \ref{sec:experiment} for details.

\noindent\textbf{Diffusion with a verbatim constraint.} We perform \emph{text diffusion}~\cite{li2022diffusion,lou2023discrete} only on stitching tokens in $\mathcal{X}$, while keeping the original context tokens frozen.
We treat the frozen tokens as observed conditioning context and perform masked denoising only on the stitching positions (i.e., text infilling~\cite{zhu2019text}). 
Let $X^{(t)}$ denote the partially denoised version of $X^{(T)}\equiv\mathcal{X}$ at reverse step $t=T,\ldots,0$. Let
$p_\theta(X^{(t-1)} \mid X^{(t)})$ denote the reverse denoising distribution
parameterized by $\theta$. Let 
$\mathcal{I}$ denote the index set of stitching
tokens in  $\mathcal{X}$. We first sample an intermediate state
\begin{equation}
\tilde{X}^{(t-1)} \sim p_\theta(X^{(t-1)} \mid X^{(t)}),
\end{equation}
and update the context state at index $i$ or $X_i$:
\begin{equation}
X^{(t-1)}_i =\begin{cases}
\tilde{X}^{(t-1)}_i, & i \in \mathcal{I},\\
X^{(T)}_i, & i \not\in \mathcal{I},
\end{cases}
\label{eq:diffusion_clamped}
\end{equation}
Our diffusion with the \emph{verbatim constraint}, i.e., denoising only stitching tokens and keeping the original context fixed, prevents factual drift while allowing coherent transitions between fragmented passages. We use pre-trained denoising models, e.g., \cite{nie2025large,ye2025dream}, as $p_\theta$.

\noindent\textbf{Why Diffusion?} Text diffusion is advantageous for context stitching for the following reasons. 1) Auto-regressive models generate tokens only from left context, i.e., $P(x_t \mid x_{<t})$, whereas stitching is an \emph{interpolation} problem capturing both preceding and succeeding contexts. We therefore adopt text diffusion, whose bidirectional attention works conditionally on both sides of the context. 2) Text diffusion enables \emph{parallel decoding} over masked tokens, which can reduce latency when the diffusion step count is much smaller than the context length.

\begin{table*}[t!]
\centering
\caption{\textbf{Comparison of retrieval-augmented generation methods. (\cmark: Retrieval used, \xmark: No retrieval).} \textbf{Bold} indicates the best performance. and \underline{Underline} denotes the second-best performance.}
\begin{adjustbox}{width=.9\textwidth}
\begin{tabular}{clcccccccccc}
\toprule
\multirow{2}{*}{\textbf{Ret.}} & \multirow{2}{*}{\textbf{Method}} 
& \multicolumn{2}{c}{\textbf{HotpotQA}} 
& \multicolumn{2}{c}{\textbf{NQ}} 
& \multicolumn{2}{c}{\textbf{TriviaQA}} 
& \multicolumn{2}{c}{\textbf{PopQA}} 
& \multicolumn{2}{c}{\textbf{2Wiki}} \\
\cmidrule(lr){3-4} \cmidrule(lr){5-6} \cmidrule(lr){7-8} \cmidrule(lr){9-10} \cmidrule(lr){11-12}
& & \textbf{EM} & \textbf{F1} & \textbf{EM} & \textbf{F1} & \textbf{EM} & \textbf{F1} & \textbf{EM} & \textbf{F1} & \textbf{EM} & \textbf{F1} \\
\midrule

\rowcolor{gray!15} \multicolumn{12}{l}{\textit{Baselines}} \\
\xmark & Naive Gen  & 16.8 & 23.8 & 17.4 & 26.2 & 56.6 & 65.5 & 19.2 & 23.3 & 8.6  & 16.5 \\
\xmark & MAD        & 23.0 & 32.7 & 21.8 & 33.1 & 56.4 & 66.3 & 21.4 & 28.2 & 18.2 & 25.1 \\
\cmark & RAG & 30.2 & 41.1 & 37.6 & 50.4 & 62.0 & \underline{71.0} & 37.0 & 45.6 & 19.0 & 28.2 \\
\midrule

\rowcolor{gray!15} \multicolumn{12}{l}{\textit{Compression Methods}} \\
\cmark & LongLLMLingua     & 29.4 & 39.5 & 33.4 & 45.4 & 60.2 & 69.3 & 34.0 & 41.5 & 16.8 & 27.0 \\
\cmark & RECOMP            & 29.2 & 39.2 & 35.2 & 46.0 & 58.3 & 65.8 & 35.8 & 43.0 & 21.4 & 33.5 \\
\cmark & Selective Context & 22.1 & 36.4 & 32.5 & 42.6 & 57.6 & 65.9 & 29.5 & 36.8 & 13.6 & 19.5 \\
\midrule

\rowcolor{gray!15} \multicolumn{12}{l}{\textit{Multi-step Methods}} \\
\cmark & SuRe       & 19.0 & 32.3 & 34.8 & \underline{51.3} & 47.6 & 64.1 & 41.8 & \textbf{48.9} & 10.2 & 18.6 \\
\cmark & FLARE      & 16.6 & 23.7 & 19.4 & 27.6 & 53.6 & 63.0 & 21.6 & 24.3 & 9.2  & 20.1 \\
\cmark & IRCoT      & 25.2 & 34.4 & 28.6 & 37.3 & 47.2 & 54.5 & 27.0 & 33.0 & 22.8 & 31.1 \\
 \cmark & Self-RAG  & 14.8 & 28.8 & \textbf{44.0} & 51.2 & 46.4 & 58.3 & 22.0 & 34.3 & 13.0 & 26.6 \\
\cmark & DRAG       & \underline{30.8} & \underline{41.7} & 36.8 & 50.3 & \underline{60.8} & 69.9 & \textbf{38.6} & 46.1 & \textbf{28.8} & \underline{34.9} \\
\midrule

\rowcolor{gray!15} \multicolumn{12}{l}{\textit{Ours}} \\
\cmark & \textbf{StitchRAG} & \textbf{32.6} & \textbf{43.4} & \underline{39.4} & \textbf{53.3} & \textbf{63.4} & \textbf{72.8} & \underline{37.2} & \underline{46.3} & \underline{27.8} & \textbf{35.4} \\
\bottomrule
\end{tabular}
\end{adjustbox}
\label{tab:main_results}
\end{table*}

\section{Experiment}\label{sec:experiment}

\subsection{Experimental Settings}
We evaluate performance on five open-domain QA benchmarks. Following prior work~\cite{jiang2023active, hu2025removal}, we randomly sample 500 examples from each dataset using the default setup provided by FlashRAG~\cite{jin2025flashrag}.

\noindent\textbf{Dataset.}
We evaluate retrieval-augmented generation on HotpotQA~\cite{yang2018hotpotqa}, Natural Questions (NQ)~\cite{kwiatkowski2019natural}, TriviaQA~\cite{joshi2017triviaqa}, 2WikiMultiHopQA (2Wiki)~\cite{ho2020constructing}, and PopQA~\cite{mallen2023not}.
HotpotQA and 2Wiki require multi-hop reasoning across multiple passages. NQ consists of real user queries paired with Wikipedia answers. TriviaQA focuses on factoid questions with evidence from web or Wikipedia sources. PopQA is an entity-centric open-domain QA benchmark constructed from Wikidata and designed to probe long-tail factual knowledge. For each query, we retrieve a fixed number of passages and condition the generator on the retrieved context. 

\noindent\textbf{Baselines.}
We compare \textsc{StitchRAG} against (i) \textit{non-retrieval} baselines, including \textsc{Naive Gen} and \textsc{MAD}~\cite{du2024improving}, (ii) Standard RAG~\cite{lewis2020retrieval} (iii) \textit{iterative / multi-step} RAG methods that involve multiple stages of retrieval and/or generation
(\textsc{SuRe}~\cite{kim2024sure}, \textsc{FLARE}~\cite{jiang2023active}, \textsc{IRCoT}~\cite{trivedi2023interleaving}, \textsc{Self-RAG}~\cite{asai2023self}, \textsc{DRAG}~\cite{hu2025removal}), and (iv) \textit{context compression} methods (\textsc{LongLLMLingua}~\cite{jiang2024longllmlingua}, \textsc{RECOMP}~\cite{xu2023recomp}, \textsc{Selective Context}~\cite{li2023compressing}).

\noindent\textbf{Implementation details.}
The system is built on the \textsc{FlashRAG} toolkit~\cite{jin2025flashrag}. Following this setup, we use E5-base-v2~\cite{wang2022text} as the retriever and the widely adopted December 2018 Wikipedia dump as the retrieval corpus~\cite{karpukhin2020dense}. For fair comparison, all methods use the same retriever, and we report Exact Match (EM) and token-level F1 using standard evaluation protocols. For \emph{single-shot} retrieval methods (including ours and non-iterative baselines), we retrieve Top-$K{=}9$ passages once and condition the generator on the retrieved context. For baselines that \emph{explicitly iterate retrieval} (e.g., \textsc{FLARE}, \textsc{IRCoT}, \textsc{DRAG}), we follow their original settings and retrieve Top-$K{=}3$ passages per iteration for up to three retrieval--generation iterations (up to 9 passages total). The generator is \textsc{LLaMA3-8B-Instruct}~\cite{huang2024llama}. We instantiate our approach with \textsc{LLaDA}~\cite{nie2025large} as the diffusion backbone, using $T=32$ reverse diffusion steps.
We set $\bar\kappa$ to 2 for intra-passage sentences and to 7 for inter-passage sentences. For the diffusion backbone, we build on Fast-dLLM~\cite{wu2025fast} and optimize its dynamic remasking for \emph{batched, GPU-parallel} execution, which is crucial in context-stitching where multiple retrieved passages are processed simultaneously. Specifically, we remove Python-level per-sample loops via vectorized sorting, prefix-product cutoff detection, and batched scatter-based selection, preserving the same remasking policy and matching transfer decisions while achieving an average $\sim$1.3$\times$ speedup for batch size $\ge 4$ (Appendix~\ref{appendix:vectorized}). All experiments were conducted on A100 GPUs. 

\noindent\textbf{Context budget for baselines.} StitchRAG reduces the retrieved context to {72.29\%} on average of its original length across datasets. To match the retention budget, we set the {retention ratio} of other compression-based baselines and fixed-retention budget methods to {75\%} in all experiments. Table~\ref{tab:compression_ratio} in Appendix~\ref{appendix:avg_retention} provides detailed statistics. 
\subsection{Main Results}\label{sec:main_results}

Table~\ref{tab:main_results} summarizes end-to-end QA performance (EM/F1) on five open-domain benchmarks. Overall, StitchRAG achieves the strongest results, ranking first on 6 of 10 metrics and second on the remaining three, demonstrating robust performance across both single-hop and multi-hop QA. On HotpotQA, it improves over RAG~\cite{lewis2020retrieval} by +2.4 EM and +2.3 F1, and also outperforms DRAG~\cite{hu2025removal} by +1.8 EM and +1.7 F1. On NQ, although Self-RAG~\cite{asai2023self} attains the highest EM, StitchRAG achieves the best F1, improving over RAG by +2.1 F1. On TriviaQA, it sets the best EM and F1, surpassing  RAG by +1.4 EM and +1.8 F1. On PopQA, DRAG achieves the highest EM, while SuRE attains the best F1. StitchRAG ranks second on both metrics. On 2Wiki, StitchRAG achieves the best F1 while ranking second in EM. Across datasets, both StitchRAG variants consistently outperform the retrieval baselines and most multi-step and compression-based methods. This shows that our dynamic context construction and diffusion-based stitching are an effective alternative to multi-round retrieval and generic context compression.

\subsection{Analyses}\label{sec:analyses}

\begin{figure}[t]
    \centering
    \includegraphics[width=0.4\textwidth]{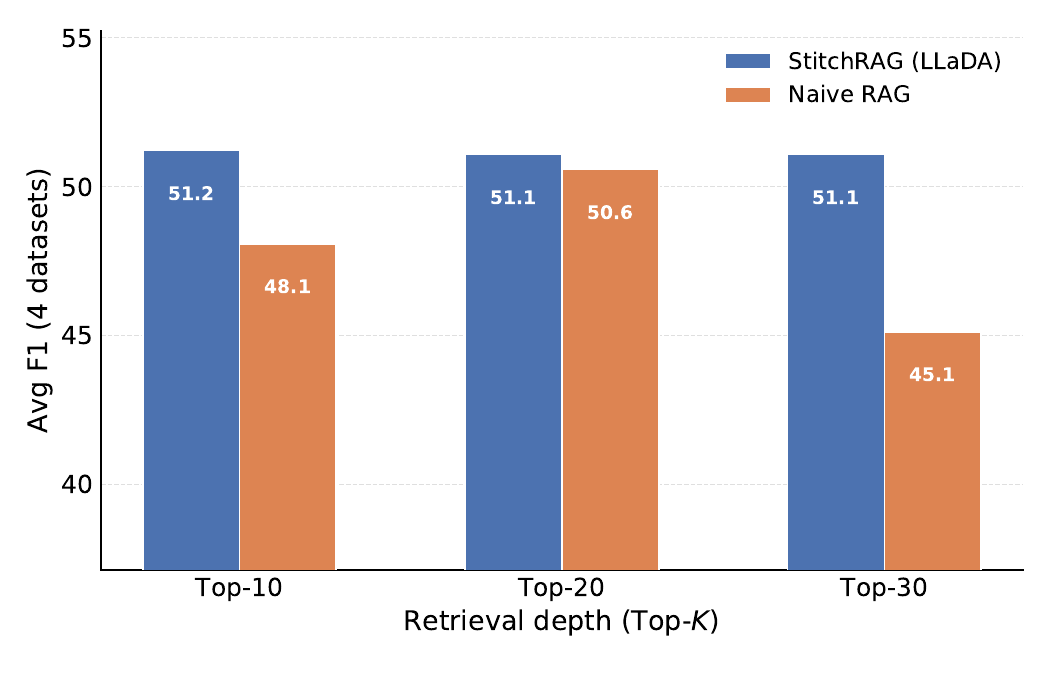}
    \caption{\textbf{Robustness to increased retrieval passages.} Average F1 across four QA benchmarks (Hotpot, NQ, TriviaQA, 2Wiki) as the retrieval number of Top-$K$ increases (Top-10/20/30). While standard RAG degrades sharply at larger $K$ due to longer and noisier contexts, StitchRAG remains stable, mitigating retrieval noise and context-length effects.}
    \label{fig:robustness_topk}
\end{figure}

\noindent\textbf{Robustness to Retrieval Noise.}
Fig.\ \ref{fig:robustness_topk} evaluates robustness under increasing retrieval depth by sweeping Top-$K$ from 10 to 30. As $K$ grows, the retrieved context becomes longer and typically noisier, substantially hurting RAG pipelines: Standard RAG drops sharply at Top-$30$, falling to 45.1 average F1 (from 48.1 at Top-$10$), indicating a pronounced sensitivity to retrieval noise and context-length effects. In contrast, StitchRAG remains essentially invariant across the sweep, staying at 51.2 average F1 for Top-$10/20/30$. This stability suggests that our dynamic context construction effectively decouples generation quality from the raw amount of retrieved evidence by filtering redundant content and preserving an information-dense context, thereby mitigating the degradation observed in naive scaling of retrieval depth.

\begin{table}[t]
\centering
\caption{\textbf{Effect of adaptive vs.\ fixed retention ratio (75\%) on StitchRAG (EM/F1).}}
\label{tab:adaptive_vs_fixed}
\begin{adjustbox}{width=\linewidth}
\begin{tabular}{l cc cc cc cc}
\toprule
\textbf{Compression} &
\multicolumn{2}{c}{\textbf{HotpotQA}} &
\multicolumn{2}{c}{\textbf{NQ}} &
\multicolumn{2}{c}{\textbf{TriviaQA}} &
\multicolumn{2}{c}{\textbf{2Wiki}} \\
\cmidrule(lr){2-3}\cmidrule(lr){4-5}\cmidrule(lr){6-7}\cmidrule(lr){8-9}
 &  \textbf{EM} & \textbf{F1} & \textbf{EM} & \textbf{F1} & \textbf{EM} & \textbf{F1} & \textbf{EM} & \textbf{F1} \\
\midrule
Fixed 75\% & 30.8 & 41.9 & 37.4 & 52.4 & 61.6 & 72.2 & 26.8 & 33.3 \\
Dynamic (Avg. 72.7\%) & \textbf{32.6} & \textbf{43.4} & \textbf{39.4} & \textbf{53.3} & \textbf{63.4} & \textbf{72.8} & \textbf{27.8} & \textbf{35.4} \\
\bottomrule
\end{tabular}
\end{adjustbox}
\end{table}

\noindent\textbf{Effectiveness of Dynamic Retention Budgets.} Table~\ref{tab:adaptive_vs_fixed} compares a fixed retention ratio (75\%) against our dynamic, utility-guided budgeting (average retention $\approx$72.7\%). Across all four benchmarks, dynamic budgeting consistently improves both EM and F1, indicating that a fixed retention rate is suboptimal under heterogeneous retrieval quality and question difficulty. The gains are most pronounced on multi-hop datasets: on HotpotQA, dynamic budgeting improves +1.8 EM and +1.5 F1, and on 2Wiki, +1.0 EM and +2.1 F1. We also observe consistent improvements on single-hop settings: on NQ, F1 increases by +0.9, and on TriviaQA, +1.8 EM and +0.6 F1. These results suggest that allocating a slightly larger budget to high-utility passages while aggressively filtering low-utility ones yields a higher information density under a comparable overall compression level, outperforming fixed-budget compression.

\begin{table}[t]
\centering
\caption{\textbf{Ablation on hybrid anchor extraction (EM/F1).} We ablate reasoning-centric AR anchors and lexical TF-IDF anchors.}
\label{tab:ablation_ar_tfidf}
\begin{adjustbox}{width=\linewidth}
\begin{tabular}{l cc cc cc cc}
\toprule
\textbf{Anchors} &
\multicolumn{2}{c}{\textbf{HotpotQA}} &
\multicolumn{2}{c}{\textbf{NQ}} &
\multicolumn{2}{c}{\textbf{TriviaQA}} &
\multicolumn{2}{c}{\textbf{2Wiki}} \\
\cmidrule(lr){2-3}\cmidrule(lr){4-5}\cmidrule(lr){6-7}\cmidrule(lr){8-9}
 &  \textbf{EM} & \textbf{F1} & \textbf{EM} & \textbf{F1} & \textbf{EM} & \textbf{F1} & \textbf{EM} & \textbf{F1} \\
\midrule
AR~$\color{blue}\checkmark$ \quad TF-IDF~$\color{red}\times$    & 29.8 & 41.4 & 39.2 & 52.1 & 61.0 & 70.7 & 25.8 & 33.4 \\
AR~$\color{red}\times$ \quad TF-IDF~$\color{blue}\checkmark$     & 31.8 & 42.4 & 37.2 & 50.9 & 63.2 & 72.4 & 26.0 & 33.9 \\
AR~$\color{blue}\checkmark$ \quad TF-IDF~$\color{blue}\checkmark$  & \textbf{32.6} & \textbf{43.4} & \textbf{39.4} & \textbf{53.3} & \textbf{63.4} & \textbf{72.8} & \textbf{27.8} & \textbf{35.4} \\
\bottomrule
\end{tabular}
\end{adjustbox}
\end{table}

\noindent\textbf{Ablation Study of Anchor Protection.} Table~\ref{tab:ablation_ar_tfidf} studies the contribution of our hybrid anchor extraction by ablating reasoning-centric AR anchors and query-centric lexical TF-IDF anchors. Using either anchor set alone is consistently worse than combining both, indicating strong complementarity. With \textsc{AR-only} (TF-IDF removed), performance drops markedly on multi-hop datasets (HotpotQA: 29.8/41.4; 2Wiki: 25.8/33.4), suggesting that AR candidates alone may miss rare entities or exact lexical cues needed for precise evidence grounding. With \textsc{TF-IDF-only} (AR removed), results improve over AR-only on HotpotQA and TriviaQA (31.8/42.4 and 63.2/72.4, respectively) but degrade on NQ (37.2/50.9), where paraphrased answers benefit from semantically guided anchors. Combining \textsc{AR+TF-IDF} yields the best performance across all datasets, demonstrating that semantic anchors from AR and lexical coverage from TF-IDF have synergistic effects, yielding faithful and information-rich sentences for subsequent stitching.

\begin{table}[t]
\centering
\caption{\textbf{Ablation on stitching components (EM/F1).} ``No stitching'' removes diffusion stitching and uses only extracted evidence spans; ``AR-stitching'' replaces diffusion stitching with an autoregressive stitcher (\textsc{LLaMA}).}
\label{tab:ablation_stitching}
\begin{adjustbox}{width=\linewidth}
\begin{tabular}{l l cc cc cc cc}
\toprule
\textbf{Variant} & \textbf{Stitcher} &
\multicolumn{2}{c}{\textbf{HotpotQA}} &
\multicolumn{2}{c}{\textbf{NQ}} &
\multicolumn{2}{c}{\textbf{TriviaQA}} &
\multicolumn{2}{c}{\textbf{2Wiki}} \\
\cmidrule(lr){3-4}\cmidrule(lr){5-6}\cmidrule(lr){7-8}\cmidrule(lr){9-10}
 &  & \textbf{EM} & \textbf{F1} & \textbf{EM} & \textbf{F1} & \textbf{EM} & \textbf{F1} & \textbf{EM} & \textbf{F1} \\
\midrule
No stitching  & -- & 32.2 & 42.7 & 38.0 & 50.9 & 61.4 & 71.1 & 26.0 & 33.4 \\
AR-stitching & \textsc{LLaMA} & 27.4 & 38.3 & 37.6 & 50.5 & 56.0 & 65.5 & 24.0 & 33.2 \\
StitchRAG & Diffusion & \textbf{32.6} & \textbf{43.4} & \textbf{39.4} & \textbf{53.3} & \textbf{63.4} & \textbf{72.8} & \textbf{27.8} & \textbf{35.4} \\
\bottomrule
\end{tabular}
\end{adjustbox}
\end{table}

\noindent\textbf{Impact of Stitching Strategies.} Table~\ref{tab:ablation_stitching} shows the effect of context stitching. We compare the modifications of StitchRAG: (i) \textsc{no stitching} performs only dynamic context construction without stitching; (ii) \textsc{AR stitching} performs stitching with autoregressive models instead of diffusion. 
Table~\ref{tab:ablation_stitching} shows that full StitchRAG performs the best on all benchmarks (e.g., HotpotQA +0.4 EM / +0.7 F1; NQ +1.4 EM / +2.4 F1; 2Wiki +1.8 EM / +2.0 F1) over \textsc{no stitching}, showing that stitching contributes non-trivial gains by producing coherent transitions and reducing discontinuities between retained spans. In contrast, AR-stitching underperforms drastically, suggesting that left-to-right generation is brittle for constrained context editing: it tends to drift from frozen evidence and accumulates errors when forced to connect disjoint fragments. Our results show that diffusion-based stitching is better suited for faithful context construction.

\begin{table}[t]
\centering
\small
\caption{\textbf{Sensitivity to the number of Anchors.} Left: AR candidate sweep with TF-IDF fixed to 10. Right: TF-IDF anchor sweep with AR fixed to 1. Values are Avg F1 on HotpotQA.}
\label{tab:hyper_sweep}
\begin{adjustbox}{width=.4\textwidth}
\begin{tabular}{lcc}
\toprule
Setting & Hyperparameter & Avg F1 \\
\midrule
 & 1 & 43.4 \\
AR candidates sweep & 2 & 43.3 \\
& 5 & 42.4 \\
\midrule
 & 5  & 42.4 \\
TF-IDF anchors sweep & 10 & 43.4 \\
 & 20 & 43.5 \\
\bottomrule
\end{tabular}
\end{adjustbox}
\end{table}

\noindent\textbf{Effects of the number of Answer Candidates.} From Table~\ref{tab:hyper_sweep}, we observe that the number of AR candidates has limited benefit beyond a single candidate. Using one AR candidate achieves a strong performance (43.4), and expanding the candidate set to 2 brings little improvement, while further increasing it to 5 substantially hurts performance. This trend suggests that additional AR candidates may introduce noisy or weakly relevant anchor signals. By contrast, performance consistently improves as the number of TF-IDF anchors increases from 5 to 10, and then saturates at 20. The small gain from 10 to 20 implies that a moderate number of TF-IDF anchors is sufficient to capture useful lexical evidence. Based on this trade-off, we use 1 AR candidate and 10 TF-IDF anchors.

\begin{table}[t]
\centering
\small
\caption{\textbf{Latency comparison.} We randomly sample 100 examples from HotpotQA. All experiments are conducted with FlashRAG~\cite{jin2025flashrag} on A100 GPUs, and results are reported as mean $\pm$ standard deviation.}
\label{tab:mean_comparison}
\setlength{\tabcolsep}{4pt}
\begin{adjustbox}{width=\linewidth}
\begin{tabular}{l|ccccc}
\toprule
& \textbf{RAG} & \textbf{Self-RAG} & \textbf{IRCoT} & \textbf{DRAG} &\makecell{\textbf{StitchRAG}}  \\
\midrule
\makecell{Latency\\(seconds)} & \makecell{$7.24$ \\ $\pm 2.84$}
& \makecell{$8.08$ \\ $\pm 0.56$}
& \makecell{$24.66$ \\ $\pm 9.81$}
& \makecell{$27.39$ \\ $\pm 7.97$}
& \makecell{$10.62$ \\ $\pm 4.30$} \\
\bottomrule
\end{tabular}
\end{adjustbox}
\label{tab:latency}
\end{table}

\noindent\textbf{Latency Analysis.}
Table~\ref{tab:mean_comparison} shows that StitchRAG strikes a practical balance between efficiency and performance. Although it is slower than Base RAG and Self-RAG, it remains substantially more efficient than IRCoT and DRAG, both of which require iterative multi-step reasoning. When we replaced diffusion-based stitching in StitchRAG with autoregressive stitching, latency increased by about 10.3 s. Thus, diffusion has advantages not only in performance but also in efficiency compared to AR due to its parallel decoding.
Overall, our result positions StitchRAG as a middle-ground design point that improves over simple one-pass baselines without incurring the full cost of iterative methods.

\noindent\textbf{Additional Results.} We provide additional experiments on the use of another diffusion backbone, \textsc{DREAM}~\cite{ye2025dream}, and an analysis of stitching tokens in Appendix~\ref{appendix:additional_experiments}. A qualitative analysis is provided in Appendix~\ref{appendix:qualitative}.

\section{Conclusion}\label{sec:conclusion}
In this work, we introduced \textsc{StitchRAG}, which addresses a central challenge in RAG: how to build a compact context without discarding the evidence most critical for answering. By combining passage-adaptive context-length allocation, hybrid anchor selection, and diffusion-based stitching, \textsc{StitchRAG} constructs compact yet information-dense contexts without incurring a substantial efficiency cost. Across five open-domain QA benchmarks, \textsc{StitchRAG} consistently outperforms prior multi-step and compression baselines while also offering a clear efficiency advantage. We hope this work motivates further research on adaptive and faithful context construction for long and complex retrieval settings.


\section{Limitations}
StitchRAG improves answer quality, reduces context length, and provides a theoretically grounded framework to determine the retention ratio. However, our approach still requires multiple model inferences similar to other advanced RAG methods. This incurs overhead as compared to the default baseline: single-pass RAG. 

To mitigate this overhead, we design the pipeline to be batch-friendly. The extra calls can be amortized via auto-regressive batch inference for anchor scoring/selection across many queries. We also use parallelization strategies for stitching, such as diffusion parallel decoding (multi-sample denoising executed in parallel, enhanced by our modification~\ref{appendix:vectorized}). In practice, these batching and parallel decoding techniques substantially reduce per-request overhead.

Finally, while our framework prevents factual drift by freezing verbatim anchors during diffusion, we restricted the diffusion method to lightweight stitching. This is mainly due to the limited capability of state-of-the-art discrete diffusion LLMs as of this writing. As diffusion-based generators mature, we envision an exciting direction beyond stitching: towards more aggressive yet faithful context refinement, elegantly rewriting and reorganizing non-anchor regions while keeping anchors intact, without increasing inference latency.

\bibliography{custom}

@article{liu2024lost,
  title={Lost in the middle: How language models use long contexts},
  author={Liu, Nelson F and Lin, Kevin and Hewitt, John and Paranjape, Ashwin and Bevilacqua, Michele and Petroni, Fabio and Liang, Percy},
  journal={Transactions of the Association for Computational Linguistics},
  volume={12},
  pages={157--173},
  year={2024}
}

@article{du2025context,
  title={Context length alone hurts LLM performance despite perfect retrieval},
  author={Du, Yufeng and Tian, Minyang and Ronanki, Srikanth and Rongali, Subendhu and Bodapati, Sravan and Galstyan, Aram and Wells, Azton and Schwartz, Roy and Huerta, Eliu A and Peng, Hao},
  journal={arXiv preprint arXiv:2510.05381},
  year={2025}
}

@techreport{hong2025context,
  title = {Context Rot: How Increasing Input Tokens Impacts LLM Performance},
  author = {Hong, Kelly and Troynikov, Anton and Huber, Jeff},
  year = {2025},
  month = {July},
  institution = {Chroma},
  url = {https://research.trychroma.com/context-rot},
}

@article{huang2024llama,
  title={The llama 3 herd of models},
  author={Huang, Kunal Chawla and Lakhotia, Kushal and Huang, Kyle and Chen, Lailin and Garg, Lakshya and Lavender, A and Silva, Leandro and Bell, Lee and Zhang, Lei and Guo, Liangpeng and others},
  journal={preprint},
  year={2024}
}

@inproceedings{asai2023self,
  title={Self-rag: Learning to retrieve, generate, and critique through self-reflection},
  author={Asai, Akari and Wu, Zeqiu and Wang, Yizhong and Sil, Avirup and Hajishirzi, Hannaneh},
  booktitle={The Twelfth International Conference on Learning Representations},
  year={2023}
}

@inproceedings{trivedi2023interleaving,
  title={Interleaving retrieval with chain-of-thought reasoning for knowledge-intensive multi-step questions},
  author={Trivedi, Harsh and Balasubramanian, Niranjan and Khot, Tushar and Sabharwal, Ashish},
  booktitle={Proceedings of the 61st annual meeting of the association for computational linguistics (volume 1: long papers)},
  pages={10014--10037},
  year={2023}
}

@inproceedings{jiang2023active,
  title={Active retrieval augmented generation},
  author={Jiang, Zhengbao and Xu, Frank F and Gao, Luyu and Sun, Zhiqing and Liu, Qian and Dwivedi-Yu, Jane and Yang, Yiming and Callan, Jamie and Neubig, Graham},
  booktitle={Proceedings of the 2023 conference on empirical methods in natural language processing},
  pages={7969--7992},
  year={2023}
}

@inproceedings{kim2024sure,
  title={SURE: SUMMARIZING RETRIEVALS USING ANSWER CANDIDATES FOR OPEN-DOMAIN QA OF LLMS},
  author={Kim, Jaehyung and Nam, Jaehyun and Mo, Sangwoo and Park, Jongjin and Lee, Sang Woo and Seo, Minjoon and Ha, Jung Woo and Shin, Jinwoo},
  booktitle={12th International Conference on Learning Representations, ICLR 2024},
  year={2024}
}

@inproceedings{hu2025removal,
  title={Removal of hallucination on hallucination: Debate-augmented RAG},
  author={Hu, Wentao and Zhang, Wengyu and Jiang, Yiyang and Zhang, Chen Jason and Wei, Xiaoyong and Qing, Li},
  booktitle={Proceedings of the 63rd Annual Meeting of the Association for Computational Linguistics (Volume 1: Long Papers)},
  pages={15839--15853},
  year={2025}
}

@inproceedings{jiang2024longllmlingua,
  title={Longllmlingua: Accelerating and enhancing llms in long context scenarios via prompt compression},
  author={Jiang, Huiqiang and Wu, Qianhui and Luo, Xufang and Li, Dongsheng and Lin, Chin-Yew and Yang, Yuqing and Qiu, Lili},
  booktitle={Proceedings of the 62nd Annual Meeting of the Association for Computational Linguistics (Volume 1: Long Papers)},
  pages={1658--1677},
  year={2024}
}

@article{xu2023recomp,
  title={Recomp: Improving retrieval-augmented lms with compression and selective augmentation},
  author={Xu, Fangyuan and Shi, Weijia and Choi, Eunsol},
  journal={arXiv preprint arXiv:2310.04408},
  year={2023}
}

@inproceedings{li2023compressing,
  title={Compressing context to enhance inference efficiency of large language models},
  author={Li, Yucheng and Dong, Bo and Guerin, Frank and Lin, Chenghua},
  booktitle={Proceedings of the 2023 conference on empirical methods in natural language processing},
  pages={6342--6353},
  year={2023}
}

@article{gong2022diffuseq,
  title={Diffuseq: Sequence to sequence text generation with diffusion models},
  author={Gong, Shansan and Li, Mukai and Feng, Jiangtao and Wu, Zhiyong and Kong, LingPeng},
  journal={arXiv preprint arXiv:2210.08933},
  year={2022}
}

@article{nie2025large,
  title={Large language diffusion models},
  author={Nie, Shen and Zhu, Fengqi and You, Zebin and Zhang, Xiaolu and Ou, Jingyang and Hu, Jun and Zhou, Jun and Lin, Yankai and Wen, Ji-Rong and Li, Chongxuan},
  journal={arXiv preprint arXiv:2502.09992},
  year={2025}
}

@article{ye2025dream,
  title={Dream 7b: Diffusion large language models},
  author={Ye, Jiacheng and Xie, Zhihui and Zheng, Lin and Gao, Jiahui and Wu, Zirui and Jiang, Xin and Li, Zhenguo and Kong, Lingpeng},
  journal={arXiv preprint arXiv:2508.15487},
  year={2025}
}

@article{lewis2020retrieval,
  title={Retrieval-augmented generation for knowledge-intensive nlp tasks},
  author={Lewis, Patrick and Perez, Ethan and Piktus, Aleksandra and Petroni, Fabio and Karpukhin, Vladimir and Goyal, Naman and K{\"u}ttler, Heinrich and Lewis, Mike and Yih, Wen-tau and Rockt{\"a}schel, Tim and others},
  journal={Advances in neural information processing systems},
  volume={33},
  pages={9459--9474},
  year={2020}
}

@article{li2022diffusion,
  title={Diffusion-lm improves controllable text generation},
  author={Li, Xiang and Thickstun, John and Gulrajani, Ishaan and Liang, Percy S and Hashimoto, Tatsunori B},
  journal={Advances in neural information processing systems},
  volume={35},
  pages={4328--4343},
  year={2022}
}

@inproceedings{jiang2023llmlingua,
  title={Llmlingua: Compressing prompts for accelerated inference of large language models},
  author={Jiang, Huiqiang and Wu, Qianhui and Lin, Chin-Yew and Yang, Yuqing and Qiu, Lili},
  booktitle={Proceedings of the 2023 conference on empirical methods in natural language processing},
  pages={13358--13376},
  year={2023}
}

@article{wu2025fast,
  title={Fast-dllm: Training-free acceleration of diffusion llm by enabling kv cache and parallel decoding},
  author={Wu, Chengyue and Zhang, Hao and Xue, Shuchen and Liu, Zhijian and Diao, Shizhe and Zhu, Ligeng and Luo, Ping and Han, Song and Xie, Enze},
  journal={arXiv preprint arXiv:2505.22618},
  year={2025}
}

@inproceedings{chen2017reading,
  title={Reading wikipedia to answer open-domain questions},
  author={Chen, Danqi and Fisch, Adam and Weston, Jason and Bordes, Antoine},
  booktitle={Proceedings of the 55th Annual Meeting of the Association for Computational Linguistics (Volume 1: Long Papers)},
  pages={1870--1879},
  year={2017}
}

@article{achiam2023gpt,
  title={Gpt-4 technical report},
  author={Achiam, Josh and Adler, Steven and Agarwal, Sandhini and Ahmad, Lama and Akkaya, Ilge and Aleman, Florencia Leoni and Almeida, Diogo and Altenschmidt, Janko and Altman, Sam and Anadkat, Shyamal and others},
  journal={arXiv preprint arXiv:2303.08774},
  year={2023}
}

@inproceedings{yang2018hotpotqa,
  title={HotpotQA: A dataset for diverse, explainable multi-hop question answering},
  author={Yang, Zhilin and Qi, Peng and Zhang, Saizheng and Bengio, Yoshua and Cohen, William and Salakhutdinov, Ruslan and Manning, Christopher D},
  booktitle={Proceedings of the 2018 conference on empirical methods in natural language processing},
  pages={2369--2380},
  year={2018}
}

@article{kwiatkowski2019natural,
  title={Natural questions: a benchmark for question answering research},
  author={Kwiatkowski, Tom and Palomaki, Jennimaria and Redfield, Olivia and Collins, Michael and Parikh, Ankur and Alberti, Chris and Epstein, Danielle and Polosukhin, Illia and Devlin, Jacob and Lee, Kenton and others},
  journal={Transactions of the Association for Computational Linguistics},
  volume={7},
  pages={453--466},
  year={2019},
  publisher={MIT Press One Rogers Street, Cambridge, MA 02142-1209, USA journals-info~…}
}

@inproceedings{joshi2017triviaqa,
  title={Triviaqa: A large scale distantly supervised challenge dataset for reading comprehension},
  author={Joshi, Mandar and Choi, Eunsol and Weld, Daniel S and Zettlemoyer, Luke},
  booktitle={Proceedings of the 55th Annual Meeting of the Association for Computational Linguistics (Volume 1: Long Papers)},
  pages={1601--1611},
  year={2017}
}

@inproceedings{ho2020constructing,
  title={Constructing a multi-hop qa dataset for comprehensive evaluation of reasoning steps},
  author={Ho, Xanh and Nguyen, Anh-Khoa Duong and Sugawara, Saku and Aizawa, Akiko},
  booktitle={Proceedings of the 28th International Conference on Computational Linguistics},
  pages={6609--6625},
  year={2020}
}

@article{wang2022text,
  title={Text embeddings by weakly-supervised contrastive pre-training},
  author={Wang, Liang and Yang, Nan and Huang, Xiaolong and Jiao, Binxing and Yang, Linjun and Jiang, Daxin and Majumder, Rangan and Wei, Furu},
  journal={arXiv preprint arXiv:2212.03533},
  year={2022}
}

@inproceedings{du2024improving,
  title={Improving factuality and reasoning in language models through multiagent debate},
  author={Du, Yilun and Li, Shuang and Torralba, Antonio and Tenenbaum, Joshua B and Mordatch, Igor},
  booktitle={Forty-first international conference on machine learning},
  year={2024}
}

@inproceedings{jin2025flashrag,
  title={Flashrag: A modular toolkit for efficient retrieval-augmented generation research},
  author={Jin, Jiajie and Zhu, Yutao and Dou, Zhicheng and Dong, Guanting and Yang, Xinyu and Zhang, Chenghao and Zhao, Tong and Yang, Zhao and Wen, Ji-Rong},
  booktitle={Companion Proceedings of the ACM on Web Conference 2025},
  pages={737--740},
  year={2025}
}

@inproceedings{karpukhin2020dense,
  title={Dense passage retrieval for open-domain question answering},
  author={Karpukhin, Vladimir and Oguz, Barlas and Min, Sewon and Lewis, Patrick and Wu, Ledell and Edunov, Sergey and Chen, Danqi and Yih, Wen-tau},
  booktitle={Proceedings of the 2020 conference on empirical methods in natural language processing (EMNLP)},
  pages={6769--6781},
  year={2020}
}

@inproceedings{pan2024llmlingua,
  title={Llmlingua-2: Data distillation for efficient and faithful task-agnostic prompt compression},
  author={Pan, Zhuoshi and Wu, Qianhui and Jiang, Huiqiang and Xia, Menglin and Luo, Xufang and Zhang, Jue and Lin, Qingwei and R{\"u}hle, Victor and Yang, Yuqing and Lin, Chin-Yew and others},
  booktitle={Findings of the Association for Computational Linguistics: ACL 2024},
  pages={963--981},
  year={2024}
}

@article{fang2025attentionrag,
  title={Attentionrag: Attention-guided context pruning in retrieval-augmented generation},
  author={Fang, Yixiong and Sun, Tianran and Shi, Yuling and Gu, Xiaodong},
  journal={arXiv preprint arXiv:2503.10720},
  year={2025}
}

@article{welleck2019neural,
  title={Neural text generation with unlikelihood training},
  author={Welleck, Sean and Kulikov, Ilia and Roller, Stephen and Dinan, Emily and Cho, Kyunghyun and Weston, Jason},
  journal={arXiv preprint arXiv:1908.04319},
  year={2019}
}

@article{mei2025survey,
  title={A survey of context engineering for large language models},
  author={Mei, Lingrui and Yao, Jiayu and Ge, Yuyao and Wang, Yiwei and Bi, Baolong and Cai, Yujun and Liu, Jiazhi and Li, Mingyu and Li, Zhong-Zhi and Zhang, Duzhen and others},
  journal={arXiv preprint arXiv:2507.13334},
  year={2025}
}

@article{hua2025context,
  title={Context engineering 2.0: The context of context engineering},
  author={Hua, Qishuo and Ye, Lyumanshan and Fu, Dayuan and Xiao, Yang and Cai, Xiaojie and Wu, Yunze and Lin, Jifan and Wang, Junfei and Liu, Pengfei},
  journal={arXiv preprint arXiv:2510.26493},
  year={2025}
}

@article{kalai2025language,
  title={Why language models hallucinate},
  author={Kalai, Adam Tauman and Nachum, Ofir and Vempala, Santosh S and Zhang, Edwin},
  journal={arXiv preprint arXiv:2509.04664},
  year={2025}
}

@inproceedings{kim2024qpaug,
  title={Qpaug: Question and passage augmentation for open-domain question answering of llms},
  author={Kim, Minsang and Park, Cheoneum and Baek, Seung Jun},
  booktitle={Findings of the Association for Computational Linguistics: EMNLP 2024},
  pages={9024--9042},
  year={2024}
}

@book{heaps1978information,
  title={Information retrieval: Computational and theoretical aspects},
  author={Heaps, Harold Stanley},
  year={1978},
  publisher={Academic Press, Inc.}
}

@article{bhat2025rethinking,
  title={Rethinking chunk size for long-document retrieval: A multi-dataset analysis},
  author={Bhat, Sinchana Ramakanth and Rudat, Max and Spiekermann, Jannis and Flores-Herr, Nicolas},
  journal={arXiv preprint arXiv:2505.21700},
  year={2025}
}

@inproceedings{zhong2025mix,
  title={Mix-of-granularity: Optimize the chunking granularity for retrieval-augmented generation},
  author={Zhong, Zijie and Liu, Hanwen and Cui, Xiaoya and Zhang, Xiaofan and Qin, Zengchang},
  booktitle={Proceedings of the 31st International Conference on Computational Linguistics},
  pages={5756--5774},
  year={2025}
}

@article{robertson2004understanding,
  title={Understanding inverse document frequency: on theoretical arguments for IDF},
  author={Robertson, Stephen},
  journal={Journal of documentation},
  volume={60},
  number={5},
  pages={503--520},
  year={2004},
  publisher={Emerald Group Publishing Limited}
}

@inproceedings{mallen2023not,
  title={When not to trust language models: Investigating effectiveness of parametric and non-parametric memories},
  author={Mallen, Alex and Asai, Akari and Zhong, Victor and Das, Rajarshi and Khashabi, Daniel and Hajishirzi, Hannaneh},
  booktitle={Proceedings of the 61st annual meeting of the association for computational linguistics (volume 1: Long papers)},
  pages={9802--9822},
  year={2023}
}

@article{zhu2019text,
  title={Text infilling},
  author={Zhu, Wanrong and Hu, Zhiting and Xing, Eric},
  journal={arXiv preprint arXiv:1901.00158},
  year={2019}
}

@article{lou2023discrete,
  title={Discrete diffusion modeling by estimating the ratios of the data distribution},
  author={Lou, Aaron and Meng, Chenlin and Ermon, Stefano},
  journal={arXiv preprint arXiv:2310.16834},
  year={2023}
}

@book{boyd2004convex,
  title={Convex optimization},
  author={Boyd, Stephen and Vandenberghe, Lieven},
  year={2004},
  publisher={Cambridge university press}
}

\appendix

\clearpage
\appendix
\onecolumn
\section{Appendix}
\subsection{Additional Experiments}\label{appendix:additional_experiments}

\begin{figure}[t]
    \centering
    \begin{minipage}[t]{0.5\linewidth}
        \vspace{0pt}
        \centering
        \begin{adjustbox}{width=\linewidth}
        \begin{tabular}{l cc cc cc cc}
        \toprule
        \textbf{Setting} &
        \multicolumn{2}{c}{\textbf{HotpotQA}} &
        \multicolumn{2}{c}{\textbf{NQ}} &
        \multicolumn{2}{c}{\textbf{TriviaQA}} &
        \multicolumn{2}{c}{\textbf{2Wiki}} \\
        \cmidrule(lr){2-3}\cmidrule(lr){4-5}\cmidrule(lr){6-7}\cmidrule(lr){8-9}
         & \textbf{EM} & \textbf{F1} & \textbf{EM} & \textbf{F1} & \textbf{EM} & \textbf{F1} & \textbf{EM} & \textbf{F1} \\
        \midrule
        Intra-only & 32.0 & 42.7 & 38.6 & 51.5 & 61.0 & 70.7 & 26.0 & 33.7 \\
        Inter-only & 32.4 & 43.1 & 39.2 & 53.1 & 61.2 & 70.6 & 26.6 & 34.5 \\
        Both & \textbf{32.6} & \textbf{43.4} & \textbf{39.4} & \textbf{53.3} & \textbf{63.4} & \textbf{72.8} & \textbf{27.8} & \textbf{35.4} \\
        \bottomrule
        \end{tabular}
        \end{adjustbox}
        \captionof{table}{\textbf{Ablation on stitching scope (EM/F1).} ``Intra-only'' uses stitching tokens only within the same passage; ``Inter-only'' stitches only across passages; ``Both'' enables both intra- and inter-passage stitching.}
        \label{tab:ablation_intra_inter}
    \end{minipage}
    \hfill
    \begin{minipage}[t]{0.47\linewidth}
        \vspace{0pt}
        \centering
        \includegraphics[width=\linewidth]{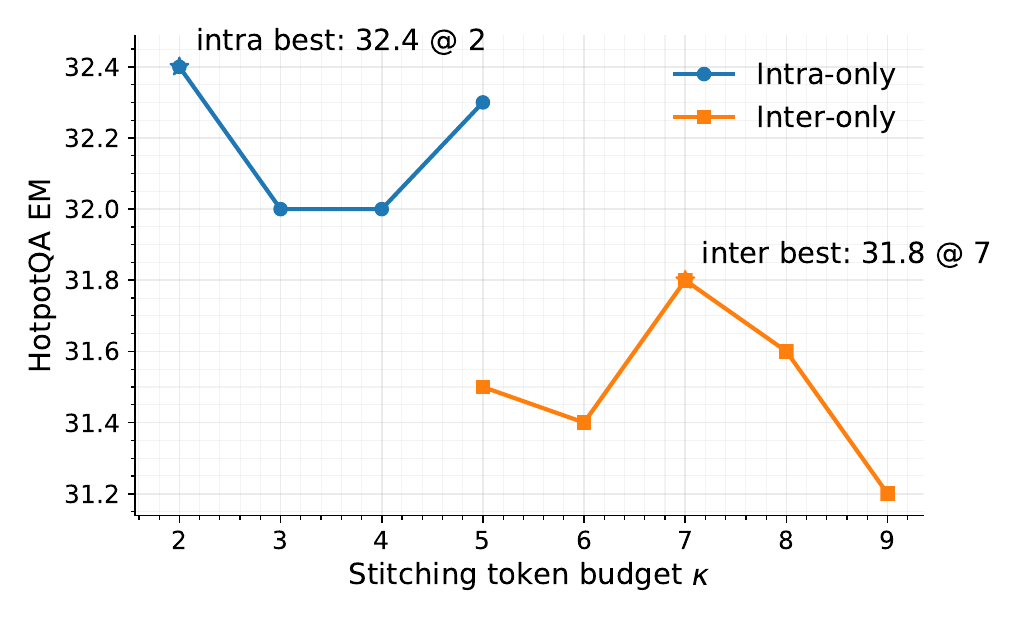}
        \captionof{figure}{\textbf{Effect of stitching-token budget on intra- vs.\ inter-passage stitching.}} 
        \label{fig:kappa_sweep_hotpot_em}
    \end{minipage}
\end{figure}

\paragraph{Analysis of Stitching token budgets.}\label{appendix:stitching_token} Figure~\ref{fig:kappa_sweep_hotpot_em} sweeps the stitching-token budget $\bar\kappa$ on HotpotQA (EM) while isolating intra-passage vs.\ inter-passage stitching. We find that the optimal budget differs substantially by scope: \textbf{Intra-only} peaks at a small budget ($\bar\kappa{=}2$, 32.4 EM), suggesting that within-passage gaps primarily require short local smoothing. In contrast, \textbf{Inter-only} benefits from a larger budget and peaks later ($\bar\kappa{=}7$, 31.8 EM), consistent with the higher difficulty of bridging cross-passage evidence. Performance degrades when $\bar\kappa$ is overly large (e.g., inter-only at $\bar\kappa{=}9$), indicating that excessive stitching capacity can introduce drift or dilute the evidence-to-stitching ratio. Overall, this motivates using distinct budgets for intra- and inter-passage stitching rather than a single global setting. 

Table~\ref{tab:ablation_intra_inter} analyzes where stitching tokens matter by restricting stitching to intra-passage gaps only or inter-passage gaps only. Using either scope alone is insufficient: \textsc{Intra-only} yields strong HotpotQA performance but lags on NQ and 2Wiki, while \textsc{Inter-only} improves cross-passage reasoning metrics but is slightly worse on HotpotQA and does not help TriviaQA. Enabling \textsc{Both} consistently delivers the best results across all datasets. This indicates that effective context construction requires \emph{both} local smoothing within retrieved passages and global bridging across passages: intra-passage stitching resolves fragment boundaries and improves readability, while inter-passage stitching is essential for multi-hop evidence composition and cross-passage coherence.

\begin{table*}[t!]
\centering
\caption{\textbf{Results of StitchRAG with different diffusion LLMs.} \textbf{Bold} and \underline{Underline} respectively denote the best performance and the second-best performance in the Table~\ref{tab:main_results} (main results).}
\begin{adjustbox}{width=.7\textwidth}
\begin{tabular}{clccccccccc}
\toprule
\multirow{2}{*}{\textbf{Ret.}} & \multirow{2}{*}{\textbf{Method}} 
& \multicolumn{2}{c}{\textbf{HotpotQA}} 
& \multicolumn{2}{c}{\textbf{NQ}} 
& \multicolumn{2}{c}{\textbf{TriviaQA}} 
& \multicolumn{2}{c}{\textbf{2Wiki}} \\
\cmidrule(lr){3-4} \cmidrule(lr){5-6} \cmidrule(lr){7-8} \cmidrule(lr){9-10} 
& & \textbf{EM} & \textbf{F1} & \textbf{EM} & \textbf{F1} & \textbf{EM} & \textbf{F1} & \textbf{EM} & \textbf{F1} \\
\midrule

\cmark & \textbf{StitchRAG (Dream)} & \underline{32.0} & \underline{42.2} & 38.6 & \underline{52.6} & \underline{62.7} & \underline{71.9}  & 27.2 & 34.7 \\
\cmark & \textbf{StitchRAG (LLaDA)} & \textbf{32.6} & \textbf{43.4} & \underline{39.4} & \textbf{53.3} & \textbf{63.4} & \textbf{72.8} & \underline{27.8} & \textbf{35.4} \\
\bottomrule
\end{tabular}
\end{adjustbox}
\label{tab:dream_results}
\end{table*}

\paragraph{Results of StitchRAG with different Diffusion Backbone (Dream).}\label{appendix:dream_results} Table~\ref{tab:dream_results} shows the performance of StitchRAG comparing different diffusion backbones for stitching: LLaDA~\cite{nie2025large} and Dream~\cite{ye2025dream}. The Dream-based variant achieves competitive performance across all benchmarks, ranking second on HotpotQA (EM/F1), NQ F1, and TriviaQA (EM/F1) compared to the baseline methods shown in the main results in Table~\ref{tab:main_results}. These results indicate that the proposed framework generalizes across diffusion backbones.

\subsection{Qualitative Analyses}\label{appendix:qualitative}

We present three qualitative case studies, each organized around a different baseline:
NaiveRAG, AR-RAG, and No Stitching.
Each case highlights how \textsc{StitchRAG} improves context construction and downstream answer generation under a distinct retrieval failure pattern.

\bigskip

\begin{tcolorbox}[colback=gray!5,colframe=black,title=Case Study: Evidence Stitching for Multi-hop Reasoning (HotpotQA)]

\textbf{Question}

\textit{When was the novel published in 2010 which is part of the series of The Maze Runner?}

\medskip
\textbf{Correct Answer}

\textit{September 18, 2010}

\medskip
\textbf{Reasoning Requirement}

\begin{center}
\small
The Maze Runner series $\rightarrow$ novel released in 2010 \\
$\Downarrow$ \\
\textit{The Scorch Trials} \\
$\Downarrow$ \\
publication date $\rightarrow$ September 18, 2010
\end{center}

\medskip
\textbf{Evidence in the Original Retrieved Context}

The retrieved passages already contain the necessary evidence, but the relevant facts are \textbf{fragmented across different sentences}.
In the original context, the two pieces of evidence are separated by
multiple sentences:

\begin{center}
\small
\textellipsis

\textbf{The Scorch Trials} is a 2010 young adult post-apocalyptic \ldots in the 
\textbf{``The Maze Runner'' series}.

The novel was published on \textbf{September 18, 2010} by Delacorte Press.

\textellipsis
\end{center}

\medskip
\textbf{Failure of Naive Context Concatenation}

The naive RAG concatenates retrieved passages into a long context
(\textbf{1607 tokens}) without explicitly connecting related evidence.
As a result, the relationship between the \textit{Maze Runner} series,
the 2010 novel, and its publication date remains implicit.
The model therefore stops at the intermediate entity:

\begin{center}
NaiveRAG $\rightarrow$ \textbf{The Scorch Trials} (incorrect)
\end{center}
\textbf{Evidence Stitching by Diffusion}

StitchRAG preserves the original evidence spans as \textit{anchors}
and generates a short connective sentence that stitches them into a
coherent reasoning chain:

\begin{center}
\small
\textellipsis

\textbf{The Scorch Trials} is the second book released \textbf{in the series}, on \textbf{September 18, 2010}.

\textellipsis
\end{center}

Importantly, the stitched sentence does not introduce new factual
evidence. Instead, it connects the preserved evidence spans into a
form that is easier for the language model to interpret.

\medskip
\textbf{Outcome}

\begin{center}
NaiveRAG $\rightarrow$ The Scorch Trials (incorrect) \\
StitchRAG $\rightarrow$ \textbf{September 18, 2010} (correct)
\end{center}

\medskip
\textbf{Context Reduction}

\begin{center}
1607 tokens $\rightarrow$ 1219 tokens \quad (24\% compression)
\end{center}

\medskip
\textbf{Analysis}

This example highlights the role of diffusion-based stitching. Although the retrieved
context already contains the necessary evidence, the relevant information appears in
fragmented sentences. StitchRAG preserves the original evidence spans and generates a
short bridging sentence that connects them into a coherent reasoning path. This stitched
context makes the relationship between evidence explicit and shortens the multi-hop
reasoning process, enabling the model to produce the correct answer.

\end{tcolorbox}

\begin{tcolorbox}[colback=gray!5,colframe=black,title=Case Study: Evidence Stitching for Multi-hop Reasoning (WikiMultiHopQA)]

\textbf{Question}

\textit{Who is James Roosevelt Roosevelt's sibling-in-law?}

\medskip
\textbf{Correct Answer}

\textit{Eleanor Roosevelt}

\medskip
\textbf{Reasoning Requirement}

\begin{center}
\small
James Roosevelt Roosevelt \\
$\Downarrow$ (half-brother) \\
Franklin Delano Roosevelt \\
$\Downarrow$ (married) \\
Eleanor Roosevelt
\end{center}

\medskip
\textbf{Evidence in the Original Retrieved Context}

The retrieved passages contain the necessary facts, but they appear in
\textbf{separate documents}. The relationship chain required for
multi-hop reasoning is therefore fragmented across the context.

\begin{center}
\small

\textellipsis

``James Roosevelt ``Rosy'' Roosevelt (April 27, 1854 – May 7, 1927) was an
American diplomat, heir, and the \textbf{older half-brother of 32nd
President of the United States Franklin Delano Roosevelt}.''

\textellipsis

``In 1905, he \textbf{married his fifth cousin once removed, Eleanor
Roosevelt}.''

\textellipsis

\end{center}

These two pieces of evidence provide the required relations:

\begin{center}
\small
James Roosevelt Roosevelt $\rightarrow$ Franklin Delano Roosevelt \\
Franklin Delano Roosevelt $\rightarrow$ Eleanor Roosevelt
\end{center}

However, in the original retrieved context these facts are separated by
large amounts of unrelated biography text, making the reasoning path
implicit.

\medskip
\textbf{Failure of Naive Context Concatenation}

Naive RAG concatenates the retrieved passages without aligning the
relations across documents. As a result, the model often stops at the
intermediate entity:

\begin{center}
NaiveRAG $\rightarrow$ Franklin Delano Roosevelt (incorrect)
\end{center}

The final relational step linking Franklin Roosevelt to Eleanor
Roosevelt is not explicitly connected to the earlier evidence.

\medskip
\textbf{Evidence Stitching by Diffusion}

StitchRAG preserves the original evidence spans and places them within a
single coherent reasoning segment:

\begin{center}
\small

\textellipsis

James Roosevelt ``Rosy'' Roosevelt was the \textbf{older half-brother of
Franklin Delano Roosevelt}.

Franklin Delano Roosevelt \textbf{married Eleanor Roosevelt}.

\textellipsis

\end{center}

This stitched context makes the relational chain explicit:

\begin{center}
\small
James Roosevelt Roosevelt $\rightarrow$ Franklin Delano Roosevelt $\rightarrow$ Eleanor Roosevelt
\end{center}

\medskip
\textbf{Outcome}

\begin{center}
NaiveRAG $\rightarrow$ Franklin Delano Roosevelt (incorrect) \\
StitchRAG $\rightarrow$ \textbf{Eleanor Roosevelt} (correct)
\end{center}

\medskip
\textbf{Analysis}

The required relations already exist in the retrieved passages but appear in
separate documents and are embedded in irrelevant text. By preserving the
evidence spans and placing them in a coherent context, StitchRAG enables the
model to connect the relations and correctly infer that the sibling-in-law
of James Roosevelt Roosevelt is Eleanor Roosevelt.

\end{tcolorbox}

\begin{tcolorbox}[colback=gray!5,colframe=black,title=Case Study: Diffusion-based Knowledge Completion (HotpotQA)]

\textbf{Question}

\textit{The rapper whose debut album was titled ``Thug Misses'' has sold over how many records worldwide?}

\medskip
\textbf{Correct Answer}

\textit{2 million}

\medskip
\textbf{Reasoning Requirement}

This HotpotQA example requires bridge-type multi-hop reasoning:

\begin{center}
\small
\textit{Thug Misses} $\rightarrow$ rapper \textbf{Khia} \\
$\Downarrow$ \\
Khia's worldwide record sales $\rightarrow$ \textbf{2 million}
\end{center}

\medskip
\textbf{Missing Evidence in the Retrieved Context}

The retrieved passages correctly identify the artist associated with
the album, but they do not contain the worldwide record sales required
to answer the question.

\begin{center}
\small
\textit{...}

``\textbf{Thug Misses} is the debut album by American rapper \textbf{Khia}.''

``The album spawned the single ``My Neck, My Back (Lick It)''.''

``As of June 2007 the album had sold \textbf{611,000 copies in the US}.'' 

\textit{...}
\end{center}

Thus, the retrieved evidence provides information about the album and
its U.S. sales, but the key fact about Khia's worldwide record sales is
missing from the retrieved context.

\medskip
\textbf{Model Outputs}

Because the retrieved passages prominently contain the number
\textbf{611,000}, several RAG systems incorrectly treat this number
as the answer.

\begin{center}
NaiveRAG $\rightarrow$ 611,000 (incorrect) \\
No Stitching $\rightarrow$ 611,000 (incorrect) \\
AR-RAG $\rightarrow$ Not specified \\
StitchRAG $\rightarrow$ \textbf{2 million} (correct)
\end{center}

\medskip
\textbf{Diffusion-based Completion}

After diffusion-based context reconstruction, the refined context
contains the following passage:

\begin{center}
\small

\textit{...}

``The album was moderately successful and achieved moderate crossover chart success.''

``Khia released her album later that year, which also gained gold status.''

``\textbf{Khia sold over 2 million records worldwide}.''

\textit{...}

\end{center}

Notably, the statement about Khia's worldwide sales does not appear
in the retrieved passages.

\medskip
\textbf{Analysis}

This example illustrates that diffusion-based stitching can go beyond
simply connecting retrieved evidence spans. While the retrieved
context contains only partial information (the album and its U.S.
sales), the diffusion process introduces a missing factual link using
parametric knowledge learned during pretraining. By combining the
retrieved evidence with this generated fact, StitchRAG is able to
complete the multi-hop reasoning chain and produce the correct answer.

\end{tcolorbox}

\begin{tcolorbox}[colback=gray!5,colframe=black,title=Case Study: Structural Evidence Stitching (HotpotQA)]

\textbf{Question}

\textit{Which author wrote a play in 2006 and a book on homosexual sex in 1977?}

\medskip
\textbf{Correct Answer}

\textit{Edmund Valentine White III}

\medskip
\textbf{Reasoning Requirement}

This question requires identifying an author who satisfies two conditions:

\begin{center}
\small
book about homosexual sex (1977)
$\rightarrow$ author
$\rightarrow$ playwright (2006)
\end{center}

\medskip
\textbf{Evidence in the Original Context}

The retrieved passages mainly discuss unrelated topics such as:

\begin{itemize}
\item religious views on homosexuality
\item legal cases and legislation
\item unrelated books and films
\item historical events
\end{itemize}

Although the word \textit{homosexual sex} appears multiple times, the passages do not
contain any information linking a specific book to its author, nor do they provide
any connection to theatrical works.

\medskip
\textbf{Structural Evidence Stitching in the Stitched Context}

The stitched context organizes evidence into structured topical segments.
For example:

\begin{center}
\small
\textbf{Documentary film} \\
\textit{Gay Sex in the 70s}
\end{center}

and the literary reference:

\begin{center}
\small
\textbf{Faggots (novel)} \\
``The novel's portrayal of promiscuous sex and recreational drug use provoked controversy.''
\end{center}

In addition, the stitched context contains structural markers such as:

\begin{center}
\small
\textbf{\#\#\# References}
\end{center}

These elements group related cultural and literary references, creating a
structured context around LGBTQ-related works.

\medskip
\textbf{How Stitching Enables Reasoning}

The stitched context introduces a crucial evidence anchor:

\begin{center}
\small
\textit{Faggots (novel)}
\end{center}

This work is associated with the author \textbf{Edmund White} and is known
for its depiction of homosexual culture in the late 1970s. This anchor
satisfies the first condition in the question:

\begin{center}
\small
book about homosexual sex (1977)
\end{center}

Through structural evidence stitching, related passages about LGBTQ literature and
cultural works are placed within the same semantic region. This stitched
context allows the model to identify \textbf{Edmund White} as the author
and recognize that he is also a playwright who produced theatrical works
in 2006.

\medskip

Rather than inserting the answer directly, \textbf{StitchRAG} stitches
topically related evidence into a coherent reasoning structure, enabling
the model to recover the reasoning chain and correctly infer the answer.

\end{tcolorbox}

\begin{tcolorbox}[colback=gray!5,colframe=black,title=Error Case]
\textbf{Question}

\textit{Who is Thessalonike of Macedon's father-in-law?}

\medskip
\textbf{Correct Answer}

\textbf{Antipater}

\medskip
\textbf{Model Prediction}

\textbf{Cassander}

\medskip
\textbf{Reasoning Requirement}

Answering this question requires a two-step kinship inference:

\begin{center}
\small
Thessalonike of Macedon $\rightarrow$ wife of Cassander \\
Cassander $\rightarrow$ son of Antipater \\
$\Downarrow$ \\
father-in-law of Thessalonike $\rightarrow$ Antipater
\end{center}

\medskip
\textbf{Evidence in the Retrieved Context}

The retrieved context clearly supports that Thessalonike married
Cassander, but it does not explicitly preserve the second link needed
for the full inference: \emph{Cassander $\rightarrow$ son of Antipater}.
Thus, the context strongly supports identifying the husband, but not the
father-in-law.

\begin{quote}
\small
[\textellipsis]

``\textbf{Thessalonike} was a Macedonian princess \textellipsis\ and
\textbf{wife of Cassander}.''

[\textellipsis]

``\textbf{Cassander} \textbf{married her}. Thessalonike became queen of
Macedon \textellipsis''

[\textellipsis]
\end{quote}

\medskip
\textbf{Example of Structural Stitching}

In the refined context, the main preserved evidence anchor is:

\begin{quote}
\small
\textbf{Thessalonike of Macedon} $\rightarrow$ \textbf{wife of Cassander}
\end{quote}

This anchor provides only \textbf{partial structural Stitching}. It links
Thessalonike to her husband, but not Cassander to his father:

\begin{quote}
\small
Cassander $\rightarrow$ Antipater's son
\end{quote}

As a result, the context encourages a \textbf{relation collapse} from
\emph{father-in-law} to \emph{husband}.

\medskip
\textbf{Reasoning Failure Induced by Incomplete Structural Stitching}

With only the first kinship link preserved, the model defaults to the
nearest available male relative:

\begin{center}
\small
Thessalonike $\rightarrow$ Cassander
\end{center}

\begin{center}
Prediction $\rightarrow$ \textbf{Cassander}
\end{center}

But the correct reasoning requires an additional genealogical step:

\begin{center}
\small
Thessalonike $\rightarrow$ husband: Cassander \\
Cassander $\rightarrow$ father: Antipater \\
$\Downarrow$ \\
father-in-law $\rightarrow$ \textbf{Antipater}
\end{center}

\medskip
\textbf{Analysis}

This is a failure case of structural stitching. The refinement preserves an
anchor sufficient to recover Thessalonike's husband, but not the extra
bridge needed to resolve the father-in-law relation. The model therefore
predicts \textbf{Cassander}, a locally supported but semantically
incorrect answer. The gold answer, \textbf{Antipater}, requires a second
genealogical link that is absent from the refined context.

\end{tcolorbox}

\begin{tcolorbox}[colback=gray!5,colframe=black,title=Error Case]
\textbf{Question}

\textit{Where does the director of film Présentation ou Charlotte et son steak work at?}

\medskip
\textbf{Correct Answer}

\textbf{Cahiers du cinéma}

\medskip
\textbf{Model Prediction}

\textbf{Switzerland}

\medskip
\textbf{Reasoning Requirement}

Answering this question requires identifying the film's director and then
linking that director to his workplace:

\begin{center}
\small
Présentation ou Charlotte et son steak $\rightarrow$ directed by Éric Rohmer \\
Éric Rohmer $\rightarrow$ worked at Cahiers du cinéma \\
$\Downarrow$ \\
director's workplace $\rightarrow$ Cahiers du cinéma
\end{center}

\medskip
\textbf{Evidence in the Retrieved Context}

The retrieved context preserves the first step but not the second, and
instead highlights an irrelevant location cue tied to the film's setting.

\begin{quote}
\small
[\textellipsis]

``\textbf{Présentation ou Charlotte et son steak} is a French short drama
film written and directed by \textbf{Éric Rohmer} \textellipsis\ In a
\textbf{Swiss village} in winter, Walter is walking with Charlotte when
they meet Clara.''

[\textellipsis]
\end{quote}

The missing second-hop fact is:

\begin{quote}
\small
Éric Rohmer $\rightarrow$ worked at \textbf{Cahiers du cinéma}
\end{quote}

\medskip
\textbf{Example of Structural Stitching}

In the refined context, the system preserves the director anchor but also
highlights a competing location anchor:

\begin{quote}
\small
\textbf{Présentation ou Charlotte et son steak} $\rightarrow$ \textbf{Éric Rohmer}

\textbf{Présentation ou Charlotte et son steak} $\rightarrow$ \textbf{Swiss village}
\end{quote}

These anchors create \textbf{misleading structural stitching}, since the
workplace link is missing while an irrelevant geographical cue remains salient.

\medskip
\textbf{Reasoning Failure Induced by Misleading Structural Stitching}

The model may therefore substitute the film's setting for the director's
workplace:

\begin{center}
\small
Présentation ou Charlotte et son steak $\rightarrow$ Swiss village \\
Swiss village $\rightarrow$ Switzerland
\end{center}

\begin{center}
Prediction $\rightarrow$ \textbf{Switzerland}
\end{center}

The correct reasoning is:

\begin{center}
\small
Présentation ou Charlotte et son steak $\rightarrow$ director: Éric Rohmer \\
Éric Rohmer $\rightarrow$ workplace: Cahiers du cinéma \\
$\Downarrow$ \\
answer $\rightarrow$ \textbf{Cahiers du cinéma}
\end{center}

\medskip
\textbf{Analysis}

This example illustrates a relation-substitution error. Although the
refined context correctly links the film to \textbf{Éric Rohmer}, it does
not preserve the evidence connecting Rohmer to \textbf{Cahiers du cinéma}.
Instead, it highlights the film's Swiss setting, which encourages the
model to substitute a salient location for the director's workplace.

\end{tcolorbox}
\subsection{Prompt Templates}\label{appendix:prompts}

\noindent\textbf{Relevance probing prompt.}
We estimate $R(x,q)=P(Z{=}1)$ using a fixed binary prompt that asks whether a retrieved passage is useful for answering the query. We decode neither a rationale nor a free-form answer; instead, we only score the canonical responses \texttt{Yes} and \texttt{No}.

\begin{quote}
\small
\texttt{Question: }\(\langle q\rangle\)\\
\texttt{Passage: }\(\langle x\rangle\)\\
\texttt{Is this passage relevant to the question? Answer (Yes/No):}
\end{quote}

\noindent\textbf{Scoring.}
Let $\mathcal{L}_{\text{Yes}}$ and $\mathcal{L}_{\text{No}}$ denote the (sequence) log-probabilities of the targets \texttt{Yes} and \texttt{No} under the model given the prompt. If a target is tokenized into multiple tokens, we sum log-probabilities across its token sequence. We then compute
\begin{equation}
R(x,q)=\frac{\exp(\mathcal{L}_{\text{Yes}})}{\exp(\mathcal{L}_{\text{Yes}})+\exp(\mathcal{L}_{\text{No}})}.
\end{equation}

\noindent\textbf{Answer generation prompt (RAG).}
For all methods, the final answer is generated by an autoregressive generator conditioned on the (possibly compressed/stitched) context. We use the following instruction-style RAG prompt.

\begin{quote}
\small
\noindent\textbf{[System]} Answer the question based on the given passages. Provide a direct and accurate answer.\\
\textbf{[User]}\\
\texttt{Passage:}\\
\(\langle \text{context} \rangle\)\\
\\
\texttt{Question: }\(\langle q\rangle\)\\
\\
\texttt{Please provide the answer in the format: `The answer is: [your answer]'.}
\end{quote}

\subsection{Average Retention ratio $r^*$ of StitchRAG}\label{appendix:avg_retention}

\begin{table}[h!]
\centering
\caption{Average $r^*$ of retrieved tokens achieved by \textsc{StitchRAG} on each dataset.}
\label{tab:compression_ratio}
\begin{adjustbox}{width=.7\textwidth}
\begin{tabular}{lcccccc}
\toprule
 & \textbf{HotpotQA} & \textbf{NQ} & \textbf{TriviaQA} & \textbf{2Wiki} & \textbf{PopQA} & \textbf{Avg.} \\
\midrule
Avg. retention ratio (\%) & 74.25 & 72.83 & 76.26 & 67.44 & 70.66 & 72.29 \\
\bottomrule
\end{tabular}
\end{adjustbox}
\end{table}

\subsection{Vectorized Implementation of Dynamic Remasking}
\label{appendix:vectorized}

\begin{table}[t!]
\centering
\caption{Comparison of implementation strategies.}
\label{tab:impl_comparison}
\begin{tabular}{lcc}
\toprule
\textbf{Component} & \textbf{Original} & \textbf{Ours} \\
\midrule
Batch iteration & Python \texttt{for} loop & Vectorized \\
Threshold computation & List comprehension & \texttt{torch.arange} \\
Sorting & Per-sample (masked-only) & Batch-parallel (full, with sentinel) \\
Cutoff detection & Nested loop & Prefix-product (\texttt{cumprod}) \\
Selection & Per-sample \texttt{topk} & Batched \texttt{scatter} \\
Edge cases & Manual \texttt{if} & \texttt{torch.where} \\
\bottomrule
\end{tabular}
\end{table}

\begin{table}[t!]
\centering
\caption{Inference time comparison of \texttt{get\_transfer\_index\_dynamic} (seq\_len=2048, vocab=32k, warmup=10, repeat=50). We conducted our experiments on a single NVIDIA A100 GPU.}
\label{tab:perf_comparison}
\begin{tabular}{lrrr}
\toprule
\textbf{Batch Size} & \textbf{Original} & \textbf{Ours} & \textbf{Speedup} \\
\midrule
1  & 2.41 ms & 2.31 ms & 1.04$\times$ \\
4  & 9.25 ms & 8.49 ms & 1.09$\times$ \\
8  & 26.49 ms & 20.00 ms & 1.32$\times$ \\
16 & 44.92 ms & 36.00 ms & 1.25$\times$ \\
32 & 84.56 ms & 65.14 ms & 1.30$\times$ \\
\bottomrule
\end{tabular}
\end{table}

This appendix compares the original Fast-dLLM dynamic remasking implementation~\cite{wu2025fast} with our vectorized PyTorch version. The vectorized implementation follows the same dynamic remasking policy as the original version and produced matching transfer decisions in all of our tests, while removing Python-level per-sample loops to improve batched inference efficiency.

\paragraph{Background (Dynamic Remasking Policy).}
Let masked tokens be ranked by confidence in descending order. Conceptually, the algorithm selects the longest confidence-sorted prefix whose elements satisfy a rank-dependent threshold,
\begin{equation}
\tau_r = 1 - \frac{\alpha}{r+1},
\end{equation}
where $\alpha$ is a factor parameter (typically $\alpha=1$), while ensuring that at least one token is transferred whenever masked positions exist.

\paragraph{Original Implementation (Fast-dLLM)}
The original implementation iterates over batch elements in Python: it computes per-sample thresholds, sorts masked-token confidences, scans ranks to find the cutoff $k$, and then applies per-sample \texttt{topk} selection.

\begin{table}[t]
\centering
\caption{Original Fast-dLLM implementation (sequential per-sample loop).}
\label{lst:original}
\begin{graycodebox}
def get_transfer_index_dynamic(logits, temperature, remasking,
                                mask_index, x, num_transfer_tokens, factor=1):
    # Sample proposals and compute confidence
    logits_with_noise = add_gumbel_noise(logits, temperature)
    x0 = torch.argmax(logits_with_noise, dim=-1)
    p = F.softmax(logits.to(torch.float64), dim=-1)
    x0_p = torch.squeeze(
        torch.gather(p, dim=-1, index=torch.unsqueeze(x0, -1)), -1)

    x0 = torch.where(mask_index, x0, x)
    confidence = torch.where(mask_index, x0_p, -np.inf)

    transfer_index = torch.zeros_like(x0, dtype=torch.bool)
    num_transfer_tokens = mask_index.sum(dim=1, keepdim=True)

    # Per-sample loop (sequential processing)
    for j in range(confidence.shape[0]):
        num_tokens = int(num_transfer_tokens[j].item())
        if num_tokens == 0:
            continue

        # Compute thresholds (Python list comprehension)
        ns = list(range(1, num_tokens+1))
        es = [factor/(n+1) for n in ns]
        threshs = [1-e for e in es]
        threshs[0] = -1  # Ensure at least one transfer

        # Sort masked tokens only
        sorted_confidence = torch.sort(
            confidence[j][mask_index[j]], descending=True)[0]

        # Find cutoff rank
        for top_i in range(len(threshs)):
            if sorted_confidence[top_i] < threshs[top_i]:
                break

        # Edge case handling
        if top_i == 0 or top_i == len(threshs)-1:
            top_i += 1

        # Select top-k
        _, select_index = torch.topk(confidence[j], k=top_i)
        transfer_index[j, select_index] = True

    return x0, transfer_index
\end{graycodebox}
\end{table}

\paragraph{Our Vectorized Implementation}
We implement the same logic using tensor-native operations only. Instead of sorting only masked positions, we sort the full sequence after assigning unmasked positions a sentinel ``$-\infty$-like'' value, ensuring they are placed at the tail and do not affect the top-$k$ masked ranks. Cutoff detection and index materialization are fully vectorized across the batch.

\begin{table}[t]
\centering
\caption{Our vectorized implementation (batch-parallel on GPU).}
\label{lst:vectorized}
\begin{graycodebox}
def get_transfer_index_dynamic(logits, temperature, remasking,
                                mask_index, x, num_transfer_tokens=None,
                                factor=1.0, threshold=None):

    # Sample proposals and compute confidence
    logits_with_noise = add_gumbel_noise(logits, temperature)
    x0 = torch.argmax(logits_with_noise, dim=-1)
    p = F.softmax(logits.to(torch.float64), dim=-1)
    x0_p = torch.gather(p, dim=-1, index=x0.unsqueeze(-1)).squeeze(-1)

    x0 = torch.where(mask_index, x0, x)
    neg_inf = torch.tensor(torch.finfo(x0_p.dtype).min,
                           device=x0_p.device, dtype=x0_p.dtype)
    confidence = torch.where(mask_index, x0_p, neg_inf)

    B, L = confidence.shape
    mask_counts = mask_index.sum(dim=1).to(dtype=torch.long)

    # Early exit for empty masks
    if (mask_counts == 0).all():
        return x0, torch.zeros_like(mask_index, dtype=torch.bool)

    # Vectorized threshold computation
    ranks = torch.arange(L, device=confidence.device, dtype=torch.float64)
    thr = 1.0 - (float(factor) / (ranks + 2.0))
    thr[0] = torch.finfo(torch.float64).min  # At least one transfer

    # Batch sorting (parallel across batch dimension)
    values, idx = torch.sort(confidence, dim=1, descending=True)

    # Vectorized threshold comparison
    pass_mask = values >= thr.unsqueeze(0)  # (B, L)

    # Prefix product to find consecutive passing ranks
    prefix_ok = torch.cumprod(pass_mask.to(torch.int64), dim=1)
    k = prefix_ok.sum(dim=1).to(dtype=torch.long)  # Per-sample k

    # Vectorized edge case handling (match original boundary handling)
    k = torch.minimum(k, mask_counts)
    k = torch.where(mask_counts > 0, torch.clamp(k, min=1), k)
    k = torch.where((mask_counts > 0) & (k == mask_counts-1),
                    mask_counts, k)

    # Vectorized top-k selection via scatter
    cols = torch.arange(L, device=confidence.device).unsqueeze(0).expand(B, L)
    select_sorted = cols < k.unsqueeze(1)
    transfer_int = torch.zeros(B, L, device=confidence.device, dtype=torch.int8)
    transfer_int = transfer_int.scatter(1, idx, select_sorted.to(torch.int8))
    transfer_index = transfer_int.bool() & mask_index

    return x0, transfer_index
\end{graycodebox}
\end{table}


\paragraph{Key Optimization}
To avoid explicit scanning loops, we implement threshold-based prefix selection with a vectorized prefix-product operation over the sorted pass mask. This yields the valid prefix length efficiently and is well-suited for batch-parallel GPU execution.

\paragraph{Performance} Both versions are dominated by sorting at long sequence lengths; the practical speedup comes from eliminating Python overhead and enabling batch-parallel execution on GPU, which improves utilization for $B\ge 4$ in our setup.

\paragraph{Equivalence Validation}
The vectorized implementation follows exactly the same threshold rule and prefix selection criterion as the original version. We verified output equivalence through randomized unit tests over 1,000 cases with varying batch sizes, sequence lengths, and mask densities, and observed no accuracy difference in end-to-end evaluation on HotpotQA and NQ.


\end{document}